\documentclass[fleqn,10pt,table,xcdraw]{wlscirep}
\usepackage[utf8]{inputenc}
\usepackage[T1]{fontenc}
\usepackage[ruled, noline]{algorithm2e}
\usepackage{algpseudocode}
\usepackage{bookmark}
\usepackage{stfloats}
\usepackage{subfig}
\usepackage{floatrow}
\usepackage{multirow}
\usepackage{adjustbox}
\usepackage{xcolor}
\DeclareMathAlphabet{\mathcal}{OMS}{cmsy}{m}{n}

\title{Enhancing Thin-Film Wafer Inspection With A Multi-Sensor Array And Robot Constraint Maintenance}

\author[1,+]{N\'estor Eduardo S\'anchez-Arriaga}
\author[1,3,*,+]{Ethan Canzini}
\author[1,+]{Nathan John Espley-Plumb}
\author[1]{Michael Farnsworth}
\author[1]{Simon Pope}
\author[2]{Adrian Leyland}
\author[1]{Ashutosh Tiwari}
\affil[1]{Department of Automatic Control \& Systems Engineering, University of Sheffield, Sheffield, UK}
\affil[2]{Department of Materials Science \& Engineering, University of Sheffield, Sheffield, UK}
\affil[3]{Airbus Robotics, Airbus UK, Chester, UK}

\affil[*]{Corresponding Author: ecanzini1@sheffield.ac.uk}

\affil[+]{These authors contributed equally to this work}


\begin{abstract}
Thin-film inspection on large-area substrates in coating manufacture remains a critical parameter to ensure product quality; however, extending the inspection process precisely over a large area presents major challenges, due to the limitations of the available inspection equipment. An additional manipulation problem arises when automating the inspection process, as the silicon wafer requires movement constraints to ensure accurate measurements and to prevent damage. Furthermore, there are other increasingly important large-area industrial applications, such as Roll-to-Roll (R2R) manufacturing where coating thickness inspection introduces additional challenges. This paper presents an autonomous inspection system using a robotic manipulator with a novel learned constraint manifold to control a wafer to its calibration point, and a novel multi-sensor array with high potential for scalability into large substrate areas. We demonstrate that the manipulator can perform required motions whilst adhering to movement constraints. We further demonstrate that the sensor array can perform thickness measurements statically with an error of $<2\%$ compared to a commercial reflectometer, and through the use of a manipulator can dynamically detect angle variations $>0.5^\circ$ from the calibration point whilst monitoring the RMSE and $R^2$ over 1406 data points. These features are potentially useful for detecting displacement variations in R2R manufacturing processes.
\end{abstract}

\begin{document}

\makeatletter
\renewcommand\@seccntformat[1]{}
\makeatother

\flushbottom

\maketitle
\thispagestyle{empty}

\section*{Introduction}

In R2R manufacturing, varied inspection systems have been explored to measure the coating thickness on large substrate areas. Existing thin-film inspection systems provide opportunities to perform measurements whilst the material is processed; however, they are not robust enough to provide reliable in-process quality control \cite{r2r_review_2022}. Imaging ellipsometry is an optical technique proven to measure coating thickness in large substrates of 300mm width but shows spatial resolution issues in the central 100mm of the web, \cite{inline_imaging_2016, inline_thickness_2019}. Atomic Force Microscopy (AFM) is a physical technique that has been studied for its potential application in R2R systems but requires high-precision and significantly large equipment to position the tip of the AFM on top of the coating surface \cite{inline_atomic_2021}. Interferometry-based techniques such as wavelength scanning interferometry (WSI) \cite{defect_assess_2014} and coherence scanning interferometry (CSI) \cite{active_optical_2020} have overcome the well-known 2$\pi$ phase ambiguity, but scaling them to cover large areas would require a significant cost and space in manufacturing in addition to the technical challenges of expanding the inspection area on large substrates. Others have created a promising approach combining Hyperspectral and RGB cameras with spectroscopic reflectometry (SR) and ellipsometry using a probabilistic sensor fusion approach to create virtual mappings of the coating surfaces. Still, these techniques require an offline physical mapping of the samples and thousands of measurements to map the wafer coating surface. \cite{vision_film_2020, high_speed_film_2021}.

SR is an alternative technique that measures a single point of the coated surface and has the advantage of “seeing” through the material and performing coating thickness measurements, nevertheless, it presents local minimum limitations when using optimisation algorithms to estimate the thickness values \cite{Kim_CNN_2019}, its accuracy decreases when inspecting rough surfaces and is normally used as an offline quality assurance tool \cite{r2r_review_2022}. However, SR is still an attractive technique due to its accuracy and low cost as compared to the other techniques mentioned above. Market-available SR systems can perform multi-point in-line measurements in roll-to-roll processes but are limited by the optical loss in the reflectance splitters \cite{hamamatsu_2021}. The physical dimensions of the light sources and spectrometers required to perform the measurements also limit system expansion to cover larger inspection areas. Despite the disadvantages, researchers are working on newer approaches using machine learning methods to predict the coating thickness for varied coatings and substrates using commercially available reflectometers. The use of machine learning methods is a viable alternative to the single-point measurement disadvantage of SR systems, however, it still requires a considerable amount of training data set to enable immediate thickness measurements \cite{thickness_eval_2021}. 

In 2021, Doo-Hyun Cho confirmed the single-point SR disadvantage was still present in the scientific community and that its use for a potential in-process inspection system for large areas would require an excessive amount of points, which suggests that it is not feasible with the existing SR technologies due to size, cost constraints and unknowns in terms of data analytics \cite{high_speed_film_2021}. In 2023, S\'anchez-Arriaga \cite{spec_low_cost_2023}, presented a miniaturised lab-based reflectometer that could potentially be stacked to create a multi-sensor array with integrated light sources which can challenge the existing single-point SR limitations, and expand the inspection of large area substrates. 

Although one of the advantages of R2R processing is its high throughput and low production cost, it presents process failures associated with roll starring and displacements observed as misalignment or fluttering of flexible substrates during operation. To simulate this scenario, and understand if the sensor array can detect substrate angle variations dynamically, a robotic arm sequence is used to validate the sensor array measurement accuracy and tilt detection capabilities for potential failures in R2R processing. This mimics the use of manipulators in thin-film wafer manufacturing for improving the automation of wafer inspection from a fabrication chamber. Robotic arms are commonly used to transfer these wafers from a fabrication chamber to inspection systems \cite{ind_rob_2023}. However, robotic manipulation remains a difficult task for applications that contain constraints in the motion of the end-effector \cite{Stilman_2007}. In this work, we learn a novel representation of these constraints for maintenance purposes to ensure the wafer remains level during manipulation, as rotations in the pitch and roll directions lead to the wafer being dropped during transportation. This ensures the correct positioning of the wafer into its calibration point for wafer inspection during manufacturing and to start performing dynamic sequences to simulate R2R process scenarios.

This paper presents a novel spectrometer multi-sensor array capable of measuring thin-film thickness across the width of Si:SiO$_2$ semiconductor wafers. This has a high potential for scalability into larger areas which is a desired feature to contribute to the global manufacturing efficiency improvements required to achieve carbon reduction emissions by 2050 \cite{nist_doc_2016}. The sensor array covers a linear width of 74mm using seven sensors positioned strategically to reduce inspection gaps and detect angle variations. Root mean squared error (RMSE) values lower than 0.02, $R^2$ greater than 0.9 and thickness error measurements below 2\% were observed per sensor, which is comparable to commercially available SR systems \cite{Avantes}. Thin-film measurements are performed via the Curve Fitting Method (CFM), calculating the root mean squared error (RMSE) between the measured reflectance curve of the coated samples (SiO$_2$) and a mathematically modelled curve. The thickness estimation was performed with the Dogbox optimisation algorithm using the Python library SciPy, and a single thickness output of the sensor array was created by performing a basic sensor fusion averaging of each sensor output\cite{elmenreich_sensor_fusion}.

\section*{Methods}

\subsection*{Multi-Sensor Array}

\subsubsection*{Array Hardware Architecture} 

\begin{figure*}[t]
    \centering
  \subfloat[\label{fig:sub-ass}]{%
       \includegraphics[width=0.195\linewidth]{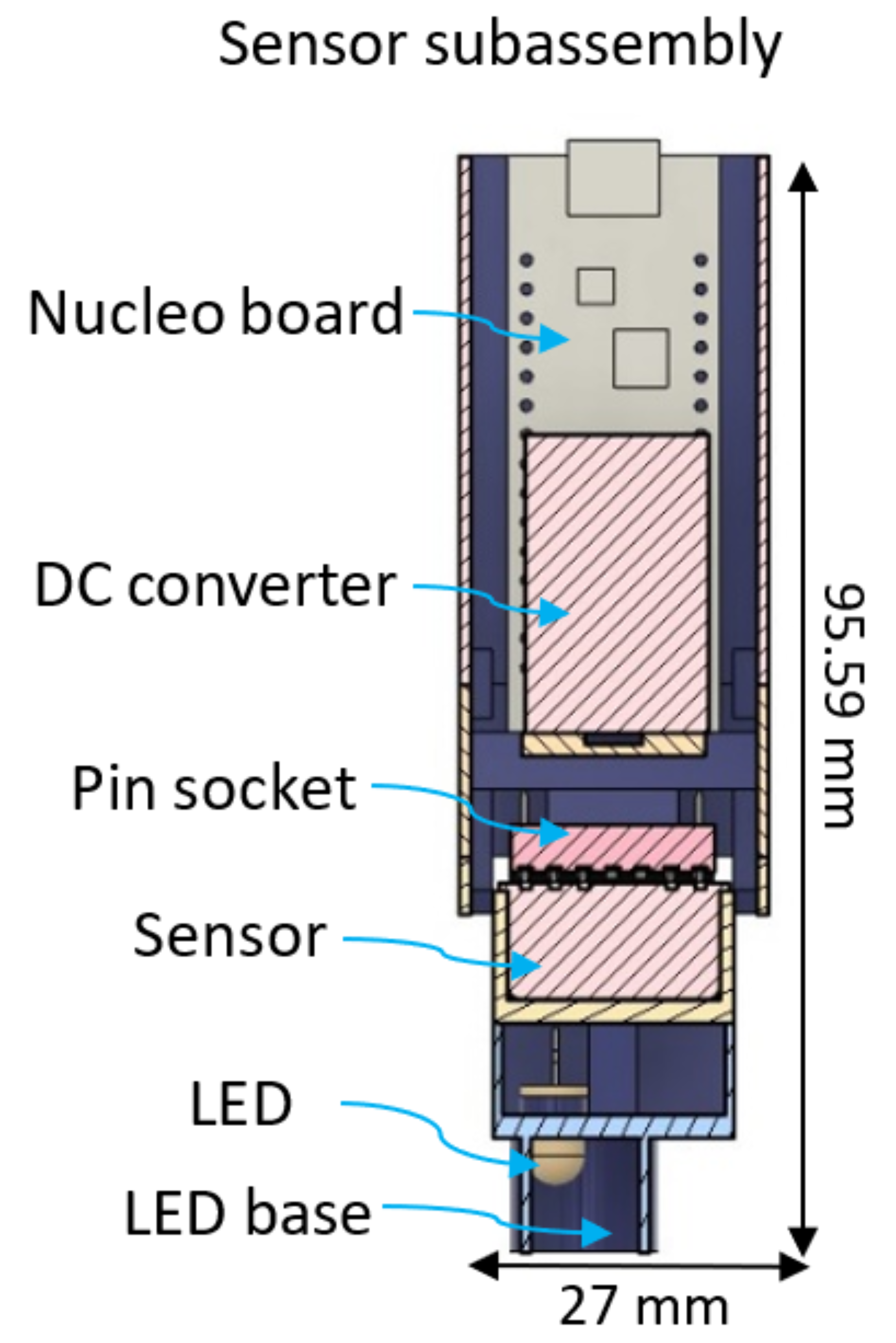}}
    \hfill
  \subfloat[\label{fig:sensor-holder}]{%
        \includegraphics[width=0.21\linewidth]{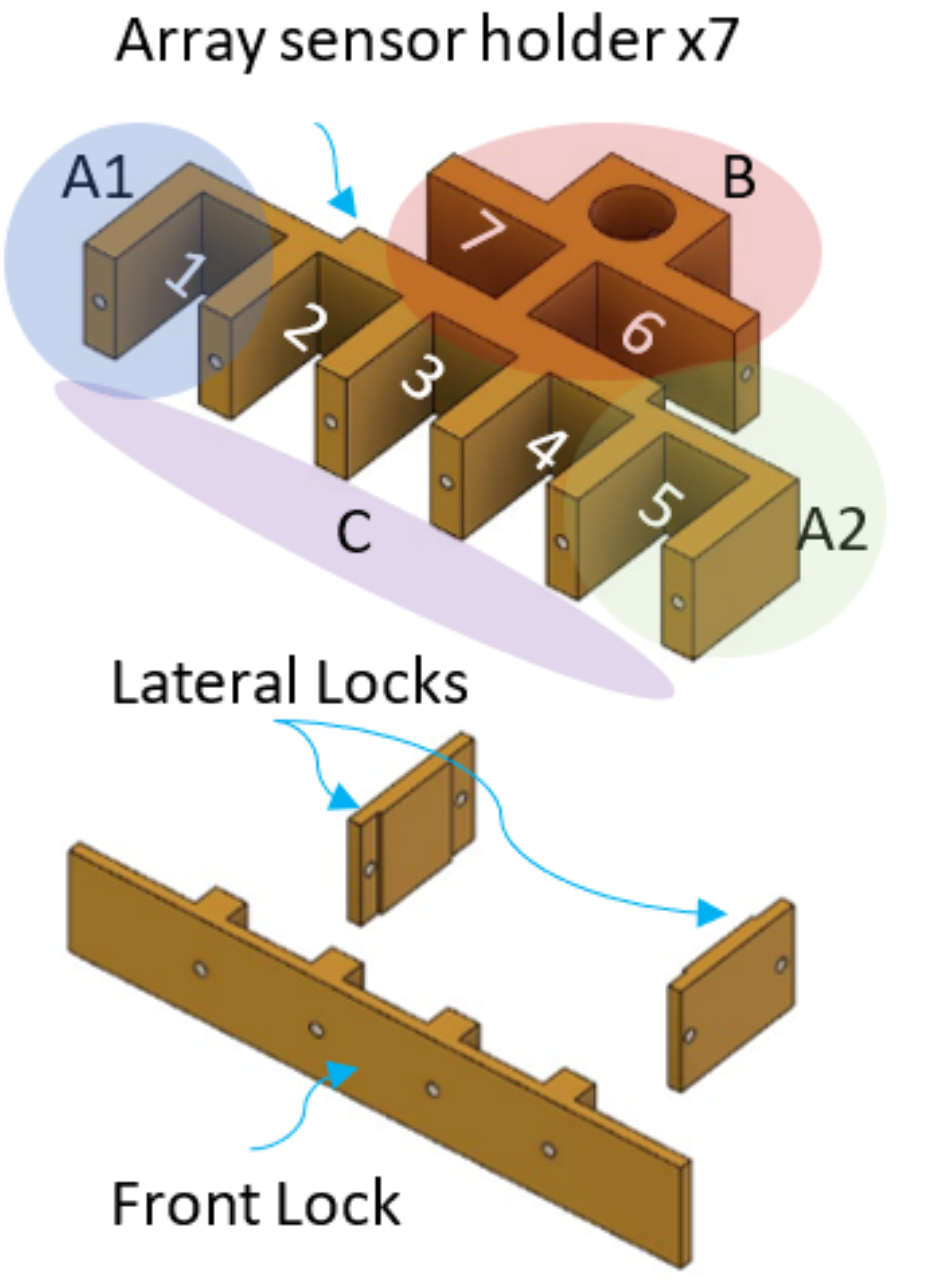}}
    \hfill
  \subfloat[\label{fig:3d-model}]{%
        \includegraphics[width=0.12\linewidth]{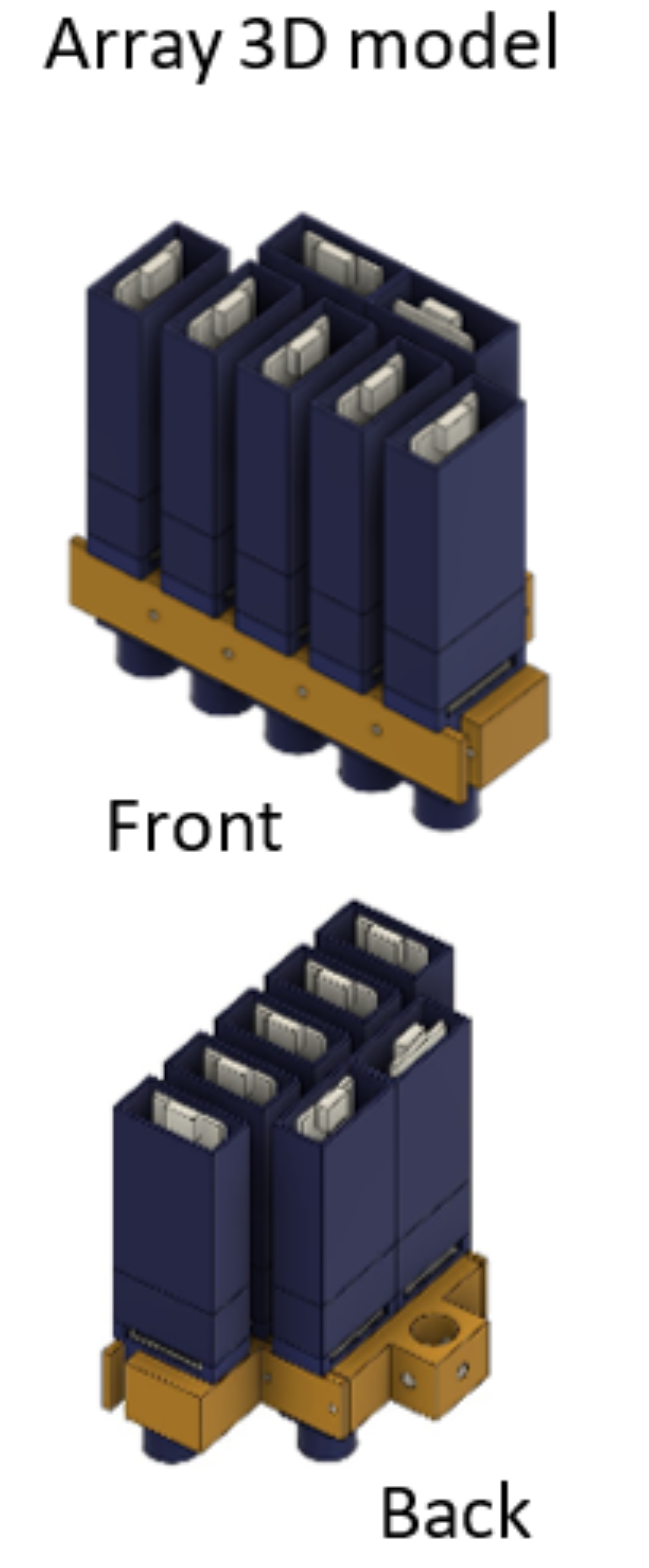}}
    \hfill
  \subfloat[\label{fig:static-model}]{%
        \includegraphics[width=0.22\linewidth]{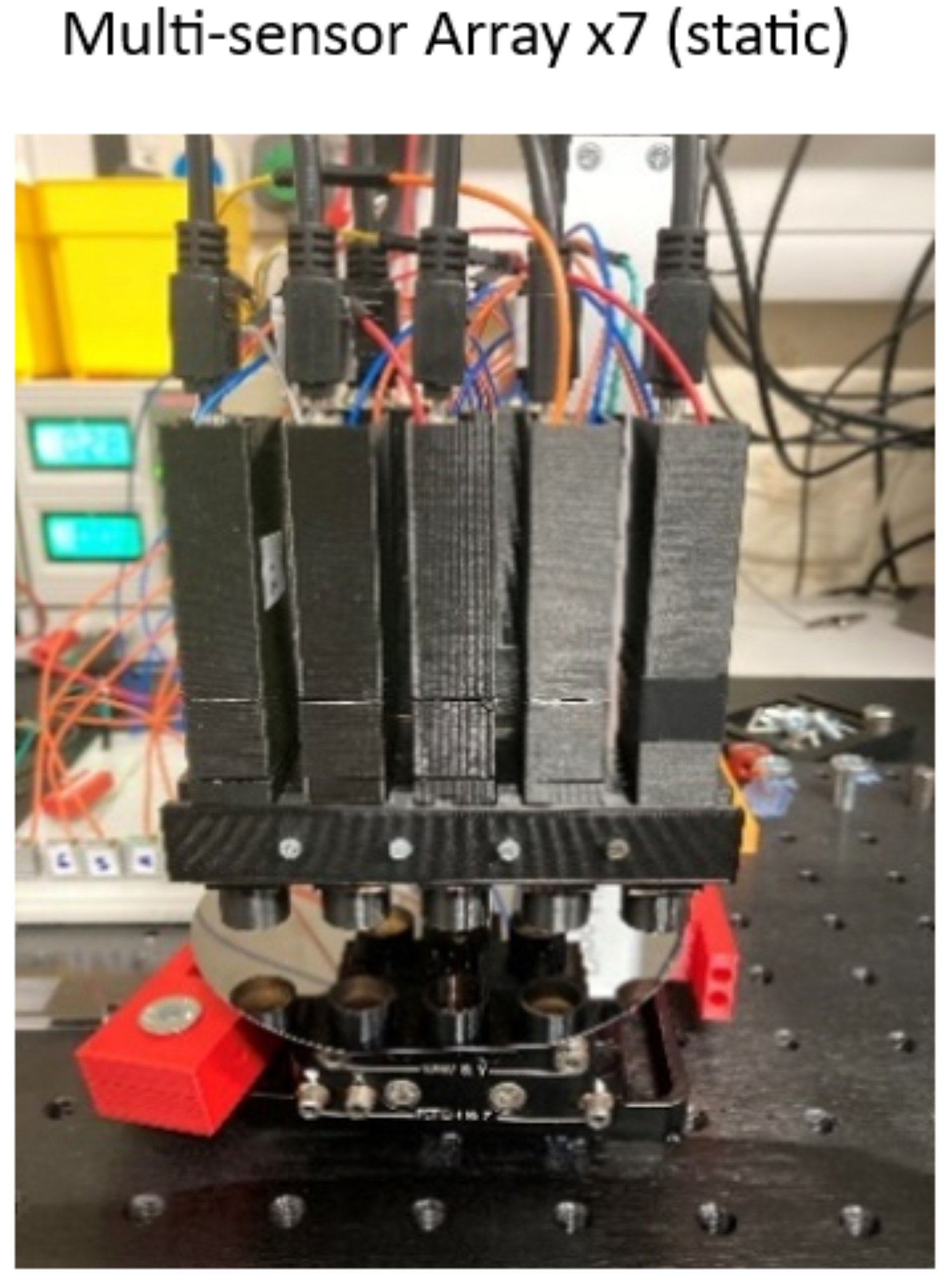}}
  \caption{Schematic of the sensor array with seven sensors: \textbf{(a)} Sensor sub-assembly cross section showing the STM Nucleo L432KC board, the DC converter, the pin socket, the sensor C12666MA and an LED; \textbf{(b)} 3D models of the array sensor holder and the lateral and front locks of the sensor devices. The top picture shows sensor zones A1-A2, B and C, and sensor positions 1-7; \textbf{(c)} Front and back isometric views of the array full assembly; \textbf{(d)} Front view of the multi-sensor array in static measurement configuration.}
  \label{fig:hardware} 
\end{figure*}

The sensor architecture is based on an original design proposed by S\'anchez-Arriaga \cite{spec_low_cost_2023}. Figure \ref{fig:sub-ass} shows the improved sensor sub-assembly designed to occupy the least space possible to a configuration of 95.59 x 27 x 13.5 mm. Figure \ref{fig:sensor-holder} shows the sensor array backbone which holds the sensor assemblies into numbered positions (1$\rightarrow$7) and strategic zones (A1, A2, B and C). Zone A1 and A2 were designed to detect left and right tilt and zone B and C to detect rear and front tilt, respectively. Sensor locks were designed to lock the sensor assemblies into a fixed position. Figure \ref{fig:3d-model} shows the front and back view of the sensor array with the sensor assemblies locked into the testing position. The locks allow an M2.5 bolt to complete the array assembly. Figure \ref{fig:static-model} shows the novel sensor array assembly held by a Dinolite microscope stand RK-10A and a sample wafer on a compact five-axis stage Thorlabs PY005/M. All sensors were connected to a DELL PC through a StarTech 7-Port Self-Powered USB-C Hub.

The sensor C12666MA is a CMOS spectrometer with 256 pixels. Each pixel corresponds to a predefined wavelength defined by the vendor as follows:  
\begin{equation}
    \lambda = A_0 + \sum_{i=1}^{5} B_i x^i
\end{equation}
\noindent where $A_0$ and $B_0 \rightarrow B_5$  are coefficients provided by the vendor and x is the pixel under study. Each pixel reads a relative intensity per wavelength in “counts,” which are defined by the microprocessor Analog-to-Digital-Converter (ADC) (max counts: $1023 = 2n - 1$, where n = 10 ADC resolution). The sensor can be enabled with an STM Nucleo-L432KC board via an Arduino-integrated development environment (IDE), with an integration time of 110 to read all pixels in 0.11 seconds and then send the full spectrum data through the COM port for data processing. 

Per manufacturing recommendation, the sensor video output must be connected to an operational amplifier (OPAMP) buffer before sending data to the microcontroller ADC, therefore, the Nucleo board L432KC was selected as it includes a configurable OPAMP in the input of its ADC. Additional power supplies were used to provide a reference voltage of 2.8V for the Nucleo board $V_{\text{REF}}$ input and another of 3.3V was used to supply the LEDs. The LED intensity was regulated externally with 2.6k$\Omega$ potentiometers to achieve 90\% of the available counts. 

\subsubsection*{Reflectance Curve Modelling}

\begin{figure}[t]
    \centering
    \includegraphics[width=0.75\textwidth]{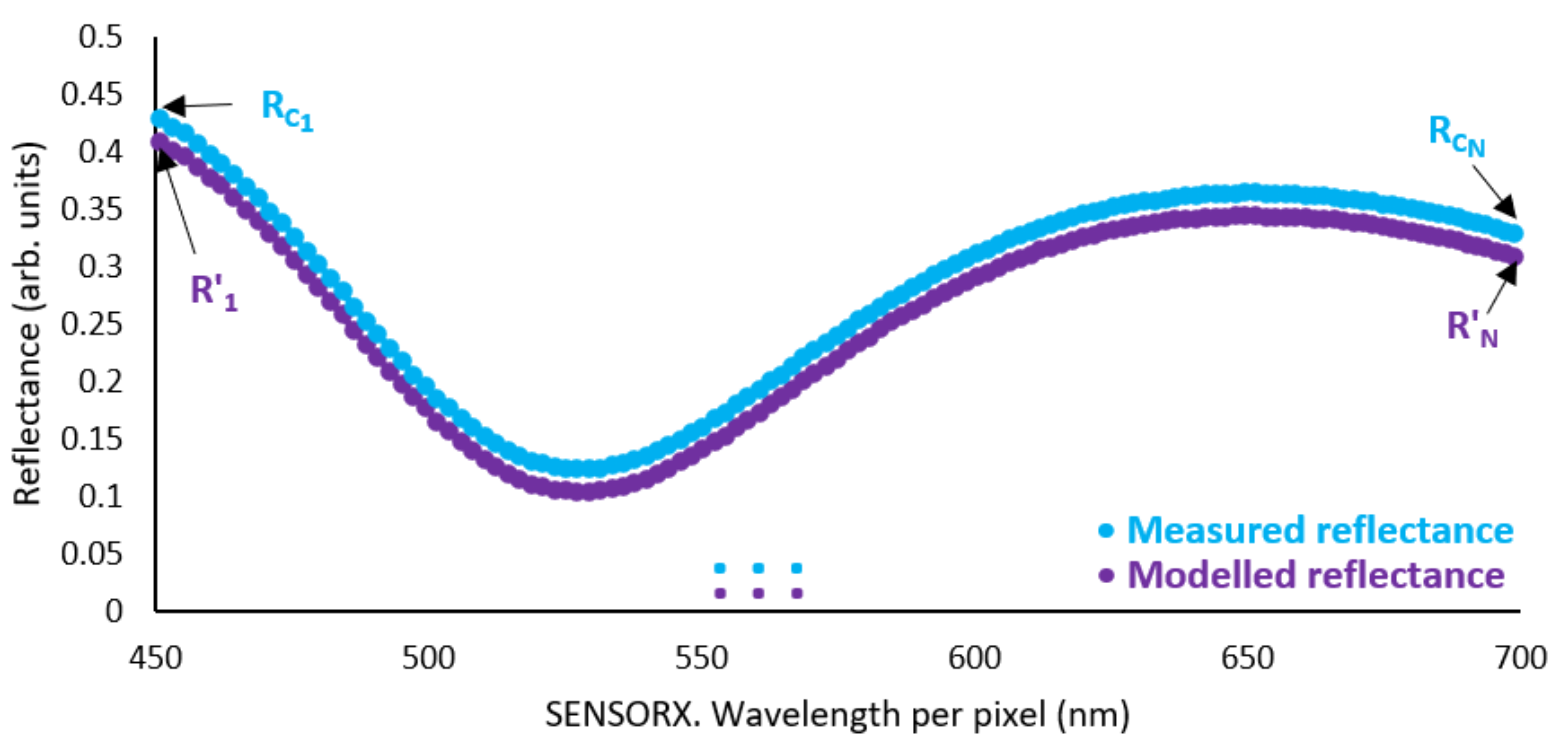}
    \caption{Reflectance curve formed by the model formula in equation \ref{eq:target} compared against the measured reflectance from equation \ref{eq:measured-reflec}.}
    \label{fig:model-curve}
\end{figure}

The reflectometer principle of operation is based on the interferometry phenomena described by Tompkins and Heavens for a single thin-film coating deposited on a semi-transparent substrate \cite{spectro_user_guide_1999, optical_film_book_1991}. When light gets reflected from an isotropic coated surface, a reflectance $R'$ data point is calculated per wavelength with:
\begin{equation} \label{eq:target}
    R' = \frac{r^2_{01}+r^2_{12}+2r_{01}r_{12}\cos2\varphi_1}{1+r^2_{01}r^2_{12}+2r_{01}r_{12}\cos2\varphi_1}
\end{equation}
\noindent where $r_{ij}$ is the total reflection coefficient per layer and $\varphi_1$ is the phase change of light in the coating. One must observe that $\varphi_1 = k'dN_1\cos\theta_1$ where $N_1$ is the coating refractive index, $d$ is the coating thickness, $\theta_1$ is the angle of incidence and $k'$ is the wave number in vacuum. In this work, $k'$ is determined as $k'=\frac{2\pi}{\lambda}$ with $\lambda$ being the wavelength under study. When a broadband light source is under study, multiple reflectance points are calculated per wavelength, resulting in a modelled reflectance curve shown in figure \ref{fig:model-curve}. The modelled reflectance curve is then compared to a measured reflectance curve per sensor. A detailed process is described in \cite{spec_low_cost_2023}

\subsubsection*{Measuring Reflectance}

Reflectance measurements require a coated sample (i.e. Si:SiO$_2$) and an uncoated sample (i.e. Si) and involves a process where three measurements must be performed. First, the uncoated sample reflected intensity ($I_U$), then the dark noise intensity ($I_D$) and finally the coated sample reflected intensity ($I_C$). Once all the intensities are measured, then the reflectance of the coated surface ($R_C$) can be calculated per pixel as follows:
\begin{equation} \label{eq:measured-reflec} 
    R_C = \frac{I_C - I_D}{I_U - I_D}R_u
\end{equation}
\noindent $R_u$ is the absolute reflectance of the uncoated sample; in practice, this value is close to one without affecting the Rc value \cite{imaging_situ_2007}. Once that Rc is measured per pixel, a reflectance curve can be generated and then compared to a modelled reflectance curve as described in the previous point of this report (see figure \ref{fig:model-curve}). The comparison between the modelled and the measured reflectance curve is performed in this work with two curve-fitting approaches. The first approach quantifies the Root Mean Squared Error (RMSE) between the modelled and the measured reflectance values. The second approach allows the thickness estimation by fitting a curve to the measured values using an optimisation algorithm. The fitted line was then evaluated using the coefficient of determination $R^2$, which allows the calculation of the RMSE as follows \cite{coefficient_2021}:
\begin{equation}
    \text{RMSE} = \frac{1}{n}\sqrt{\sum_{i=1}^{n} \left ( Y - \hat{Y} \right )^2}
\end{equation}
\noindent where $n$ is the number of pixels under comparison, $Y$ is the measured value per wavelength/pixel and $\hat{Y}$ is the modelled value per wavelength/pixel. The RMSE values must be close to zero to ensure reliable data. Since the quality metric is defined by the user \cite{Avantes, ops_manual_filmetrics}, and based on previous work, in this study, an RMSE value < 0.04 per sensor is sufficient to ensure, a reliable thickness estimation.

\subsubsection*{Coating Thickness Estimation}

The thickness estimation was performed via the Python SciPy curve fitting function using the Dogbox optimisation algorithm \cite{dogleg_1970}. This function fits a line within the measured reflectance curve per sensor and returns optimised values that best describe the measured data. Additionally, it showed the best capability performance after running a Minitab six-pack analysis compared to the other optimisation algorithms available in Python SciPy library (the well-known Levenberg-Marquardt and Trust Region Reflective (TRF)). To achieve this, we must provide a target function to optimise, equation \ref{eq:target}, and two groups of data. The first group is the x and y values which are the pixel wavelengths and the measured reflectance values per pixels. The second group is the estimated thickness, refractive index, and the angle of incidence. The second group of data is the one that the curve fitting function optimises to achieve the least error between the measured reflectance values and the fitted reflectance line.

Once the curve fitting function process is completed, the coefficient of determination $R^2$ is used to understand the goodness-of-fit between the measured and the fitted values \cite{coefficient_2021}: 
\begin{equation}
    R^2 = 1 - \frac{\sum(Y - Y_f)^2}{\sum(Y - \Bar{Y})^2} = 1 - \frac{SSR}{SST}
\end{equation}
\noindent where $Y$ is the measured value per wavelength/pixel, $Y_f$ is the fitted value per wavelength/pixel and $\Bar{Y}$ is the mean of the measured values. The numerator is also known as the sum of the squared residuals (SSR) and the denominator is also known as the total sum of squares (SST). In this work, for exploratory purposes, an $R^2$ value > 0.7 is considered a reliable value for the thickness estimation per sensor. The estimated thickness value from the curve fitting function is used as the measured thickness per sensor.

\subsubsection*{Sensor Fusion \& Noise Handling}
The definition of "sensor fusion" has been questioned for the last three decades and recently has regained controversy in academia due to the increased complexity and evolution of technology, applications, and fusion algorithms.  Nevertheless, according to the definitions by Elmenreich \cite{elmenreich_sensor_fusion} and Klein \cite{sensor_data_fusion_Klein} this paper considers "sensor fusion" as the combination of $n$ sensors to have a better representation of the wafer area thickness under inspection. According to the central limit theorem, the thickness measurements of the individual sensors should converge close to a normal distribution, which is a proven assumption for our sensor array \cite{hamamatsu_2021_technical_info}. Therefore, the sensor fusion technique was a simple averaging (SA) performed to obtain the average thickness $\Bar{Y}$ \cite{elmenreich_sensor_fusion} \cite{remote_sensing_2022}:
\begin{equation}
    \Bar{Y} = \frac{1}{n}\sum_{i=1}^{n} Y_i
\end{equation}
\noindent where $Y_i$ is the individual thickness measurement from each sensor in the array $n \in \{1, 7\}$. 

To mitigate the effect of sensor and USB noise on the readings, a convolution filter was applied to smooth the intensity values and eliminate noise.

\subsection*{Learning Constraint Manifolds in Robotics}

\subsubsection*{Constraint Manifolds}

Manifolds represent a subset of geometry dealing with curvature that exists in higher dimensions \cite{intro_manifolds_2018}. Many aspects of robotic manipulation can be considered to operate on manifolds, such as the symmetric and positive definite (SPD) matrices for joint stiffness and unit quaternion (UQ) for orientation \cite{wang_deep_robot_skills_2022}. These manifolds, denoted as $\mathcal{M}$, can be considered to be Riemannian and dictate the capabilities of specific robotic platforms. Many robotic applications require the use of custom end-effectors, such as the wafer transportation tool used in this work. Some platforms may however be restricted in what positions the end-effector can take, with tasks such as opening doors or drawers \cite{Stilman_2007} imposing end-effector constraints on rotation and position. Traditionally, these constraint manifolds are identified iteratively during the path-planning process through the sampling of joint configurations that satisfy the desired constraint:

\begin{equation} \label{eq:constraint-manifold}
\mathcal{M} := \{\theta \in \Phi\ |\ \Phi^- \preceq \theta \preceq \Phi^+ \text{ and } \mathbf{f}(\theta) = 0\}
\end{equation}

\noindent where $\Phi$ is the joint limits of the robot, $\theta$ is a set of joint positions from $\Phi$ and $\mathbf{f}(\theta)$ is the constraint function. For each new joint configuration sampled, the manifold grows and is identified during the planning stage. However, for complex constraint functions that limit movement on multiple axes, this process can increase planning time and reduce the efficiency of many algorithms. Additionally, many repetitive tasks that maintain constraints which don't change during product life cycles, meaning that manifolds need only be identified once to ensure compliance with robot movement. 

Manifold learning represents a method to identify the properties of a high-dimensional manifold through prior offline data collection. Formally, it is the process of defining a function $f$ that maps some Euclidean space into a lower dimensional manifold \cite{reactive_vae_2022}:

\begin{equation}\label{eq:manifold-func}
    \mathcal{M} = \mathbf{f}(\mathbf{x}) \text{ with } \mathbf{f}:\mathcal{X} \rightarrow \mathcal{Z}
\end{equation}

\noindent $\mathcal{X}$ represents the Euclidean space and $\mathcal{Z}$ represents a closed subset of $\mathcal{X}$ lying on $\mathcal{M}$. Determining the mapping function $\mathbf{f}$ becomes complex in robotics tasks as there is a coupling between the Euclidean motion group $SE(3)$ and the joint configuration space $Q$. Approximating the function $\mathbf{f}$ is seen as a way to avoid the "curse of dimensionality" within high-dimensional models, where the function approximator can interpolate between the data points to construct the manifold. Another benefit of using manifold learning for robotics is the ability to evaluate the relationship between the joint space $\Phi$ and the pose space. Once the joint space relationship is known, it can be used directly to evaluate when the robot experiences drift from being on-manifold and maintenance to the robot is required. 

\subsubsection*{Constraint Manifold Identification}

For the constraint function, we refer back to the original problem statement of this paper whereby we transport a wafer sample from a fabrication chamber to the sensing array for inspection. As the wafer is deposited on the end-effector of the robot and isn't locked into place, we must impose a horizontal constraint on the end effector \cite{Stilman_2007}. Consider the transformation matrix $\mathbf{T}^0_e$ that relates the pose in the end effector frame $\mathcal{F}^e$ relative to the base frame $\mathcal{F}^0$, for a robot with $n$ joints:

\begin{equation} \label{eq:transformation}
    \mathbf{T}^0_e = \mathbf{T}^0_1\ \mathbf{T}^1_2\ ...\ \mathbf{T}^{n}_e = \left ( \prod_{i=0}^{n-1}\ \mathbf{T}^{i}_{i+1} \right )\ \mathbf{T}^{n}_e = \begin{pmatrix}
 \mathbf{R}^{0}_e & \mathbf{P}^0_e\\ 
 0 & 1
\end{pmatrix}
\end{equation}

Equation \ref{eq:transformation}'s rotations and translations are determined using the forward kinematics of the robot manipulator from the joint angles $\theta$. The rotations of the robot are normally expressed in three main ways, the first being the rotation matrix $^{0}\mathbf{R}_e$ shown in equation \ref{eq:transformation}. In this work, we represent the rotation of the end effector in the RPY representation of Euler's angles $[\beta\ \alpha\ \gamma]^\intercal$ for our constraint manifold identification. This brings a binary constraint vector in the form:

\begin{equation}\label{eq:constraint}
    \mathbb{C}_{RPY} = \left [ c_x\ c_y\ c_z\ c_\beta\ c_\alpha\ c_\gamma \right ]^\intercal
\end{equation}

\noindent to constrain the 6 degrees of freedom pose of the end-effector of a manipulator operating in Euclidean space $\mathbb{R}^3 \times SE(3)$, henceforth denoted as $\mathcal{X}$ as shown in equation \ref{eq:manifold-func}.

Now that we have defined the transformation matrix $\mathbf{T}^0_e$ and the constraint array $\mathbb{C}_{RPY}$, we now can formulate our constraint function $\mathbf{f}(\theta)$. Using the joint positions $\theta$, the forward kinematics of the manipulator can be computed to find the pose of the end effector $\mathbf{x}^0_e$ relative to the base frame $\mathcal{F}^0$. Using this, the constraint function can be constructed as the $\ell_2$-norm of the element-wise product of the constraint array and pose vector:

\begin{equation}\label{eq:con-func}
    \mathbf{f}(\theta) = \lVert \mathbb{C}\odot\mathbf{x}^0_e \rVert_2 \text{ where } \mathbf{x}^0_e \equiv \mathbf{T}^0_e (\theta)
\end{equation}

Sampling the joint positions is done through a Monte Carlo sampling method, whereby increasing the number of samples can improve the overall estimation of the manifold. 

\subsubsection*{Learning Manifolds From Data}

\begin{figure}[t]
    \centering
    \includegraphics[width=0.8\linewidth]{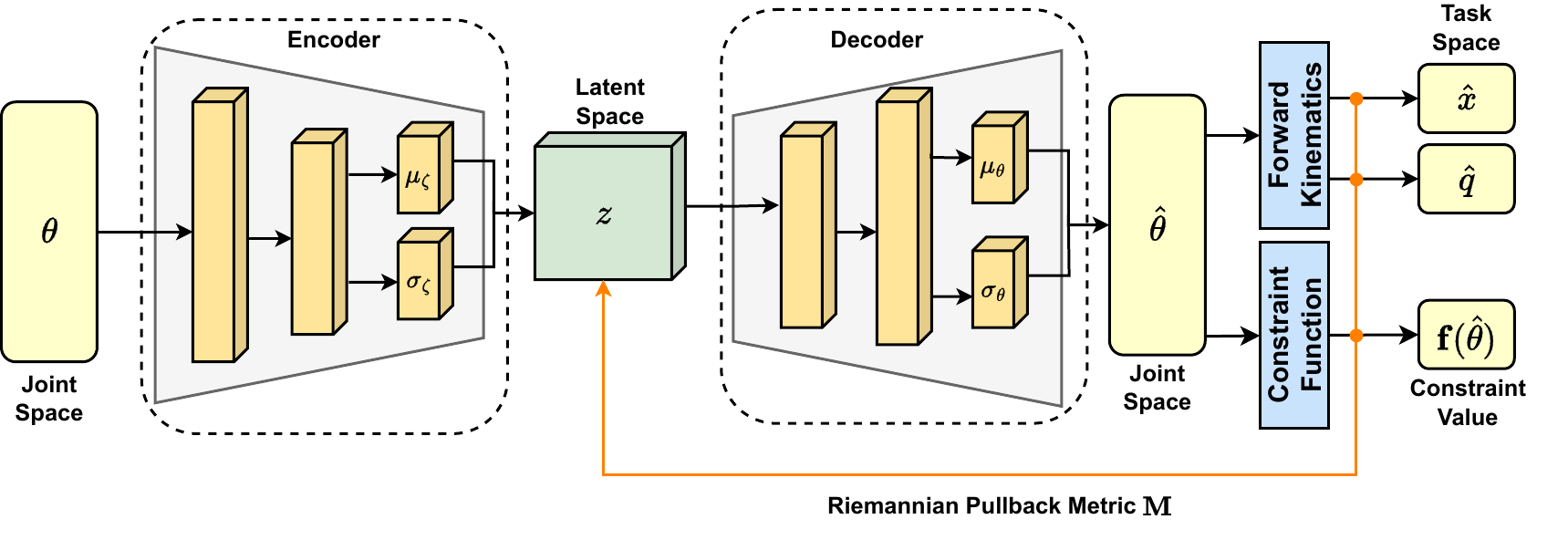}
    \caption{Model of the VAE system for generating a latent space Riemannian manifold. The latent space $q_\zeta(z | \theta)$ is used to generate the Riemannian metric $\mathbf{M}$, which is used to determine whether the manipulator is experiencing joint drift. The estimate of the constraint function $\hat{\mathbf{f}}(\theta)$ is computed from decoding the latent space and computing the manifold constraint function}
  \label{fig:vae-constraint}
\end{figure}

Learning manifolds from data relies on reducing a higher dimensional manifold into a lower dimensional space through a projection function $\phi(z)$. This projection function can be modelled as the latent space of a function that learns representations between the joint positions, whereby it learns a Riemannian manifold on this latent space. The variational autoencoder (VAE) \cite{vae_2013} learns a latent space representation of data input $\mathbf{x}$. Deep VAE models seek to maximise the evidence lower bound (ELBO) of the model, which we can modify to use the joint space of the robot manipulator to produce a lower dimensional manifold of the joint operating positions \cite{reactive_vae_2022}:
\begin{equation} \label{eq:elbo}
    \mathcal{L}_{ELBO} = \mathbb{E}_{q_{\zeta}(z | \theta)}\left [ \log(p_{\Theta}(\theta | z)) \right ] - \text{KL}\left [ q_{\zeta}(z | \theta) \parallel p_\phi(z, \theta) \right]
\end{equation}
\noindent where KL denotes the Kullback-Leibler divergence between the encoder distribution $q_\zeta(z | \theta, \mathbf{f}(\theta)$ and the Gaussian latent variables and $p_\Theta$ is the joint space conditional density. Once the model has been trained, the Riemannian metric first derived in \cite{reactive_vae_2022} can be used in combination with the predicted constraint value on the manifold $\hat{\mathbf{f}}(\theta)$ for the manipulator joint positions to generate a Riemannian metric corresponding to the joint constraint value:
\begin{equation}\label{eq:r-metric}
    \mathbf{M}^\theta_{\mathbf{f}}(z) = \zeta\left | -1 + \exp[\hat{\mathbf{f}}(\mu(z)] \right |
\end{equation}
This metric takes large values in areas where the model has a high uncertainty regarding joint positions and when estimated constraint function $\hat{\mathbf{f}}(\hat{\theta})$ takes a large value. The model architecture is shown in figure \ref{fig:vae-constraint}, which was deployed into the MoveIt planning interface so plans can be evaluated and determine whether maintenance is required for calibrating the joint positions. 

\section*{Results}

\subsection*{Wafer Inspection Results}

\subsubsection*{Static Experiments - Inspection Box Definition}

\begin{figure*}[h]
    \centering
  \subfloat[\label{fig:setup-1}]{%
       \includegraphics[width=0.25\linewidth]{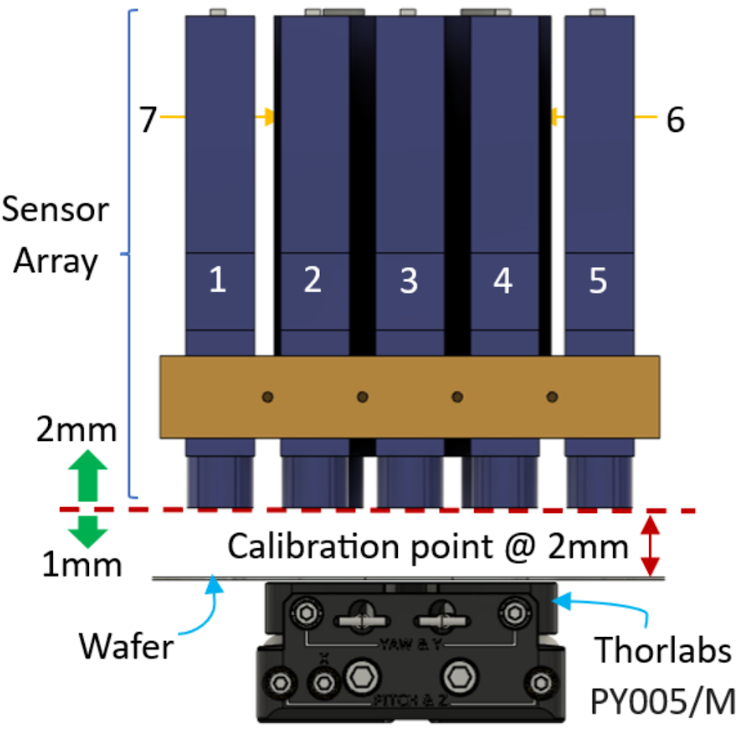}}
    \hspace{1cm}
  \subfloat[\label{fig:setup-2}]{%
        \includegraphics[width=0.27\linewidth]{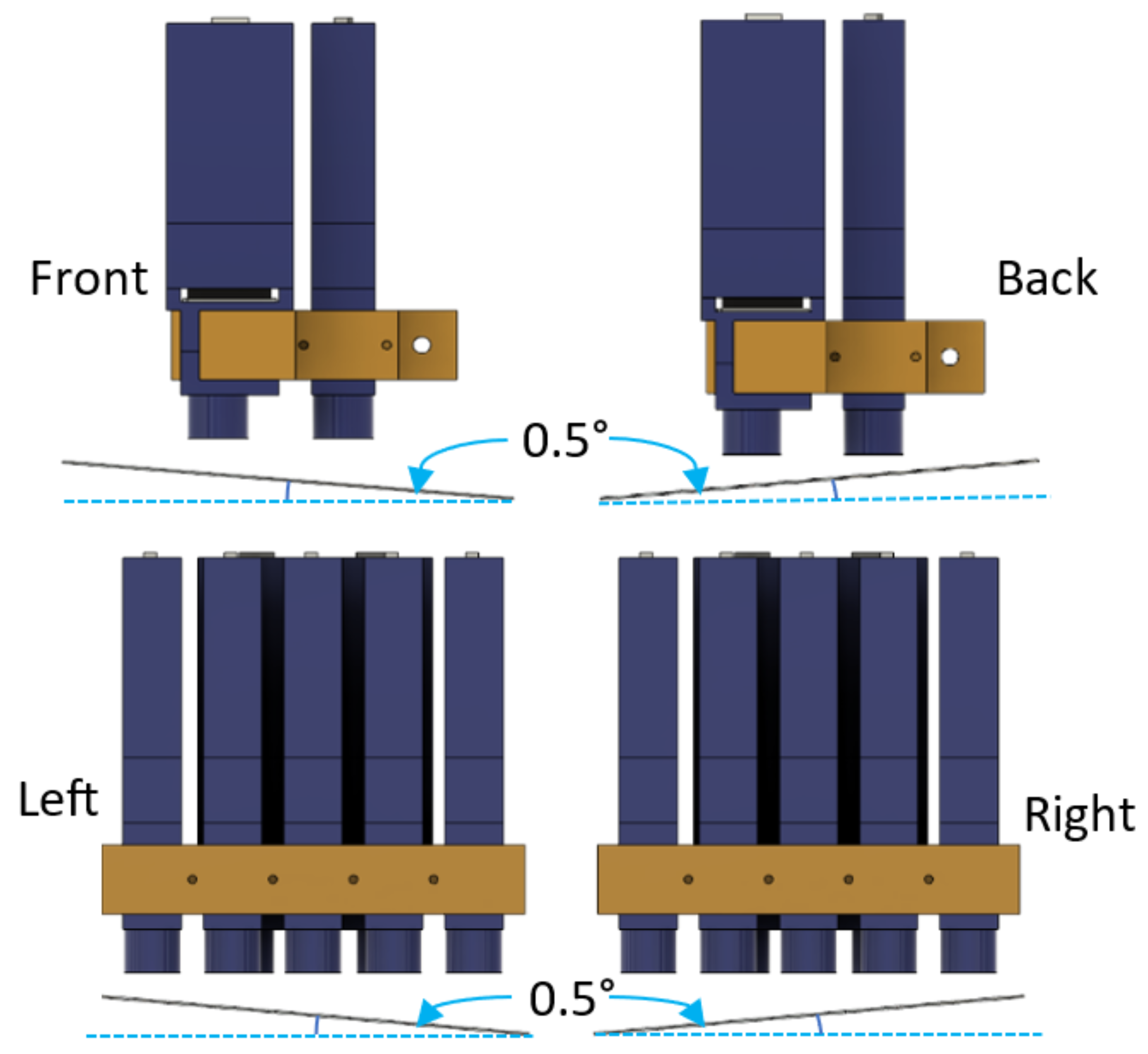}}
    \hspace{1cm}
  \subfloat[\label{fig:setup-3}]{%
        \includegraphics[width=0.2\linewidth]{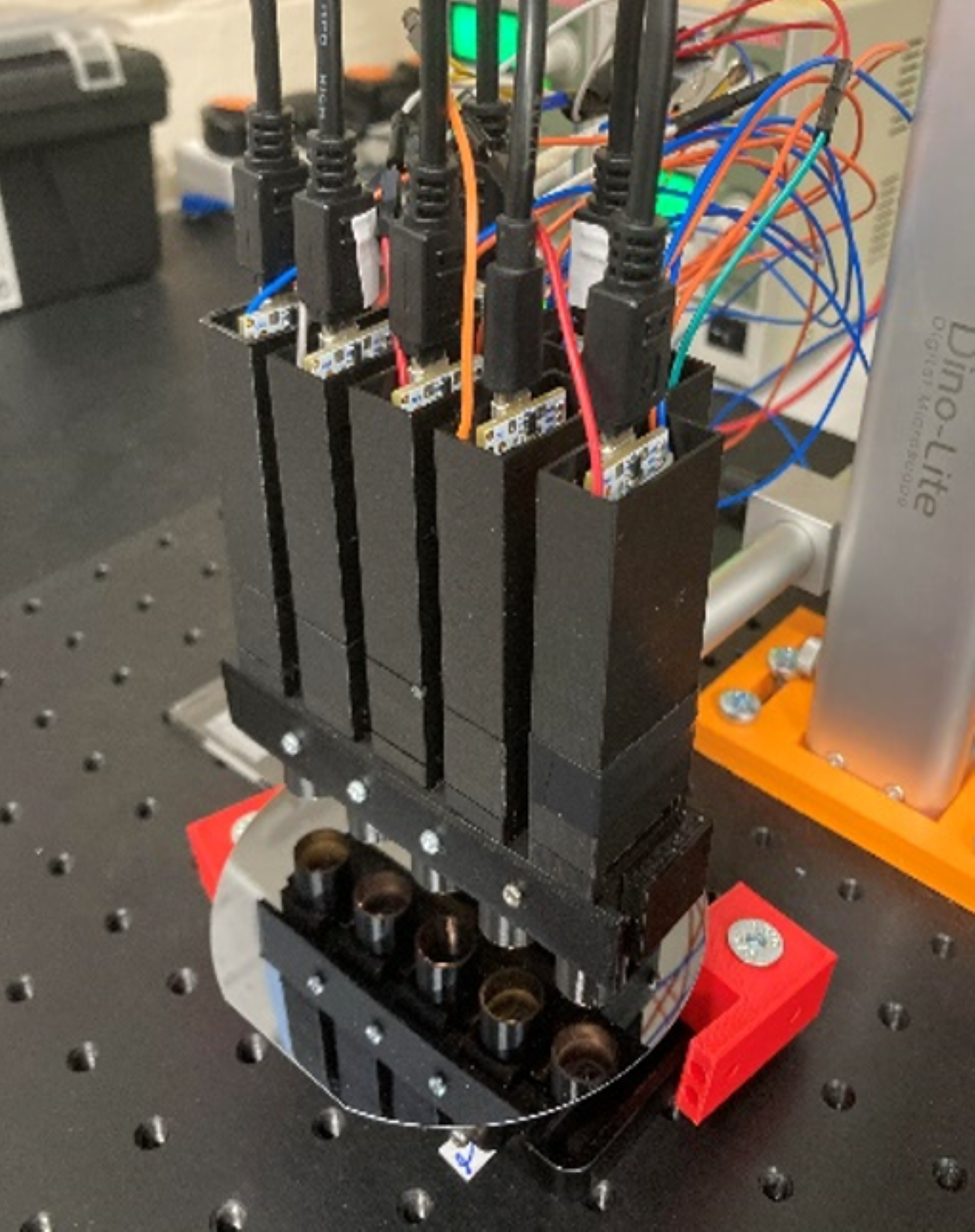}}
  \caption{Array height and angle experiments. \textbf{(a)} Sensor array calibration point at 2mm above wafer surface (red dotted line) and height variations from the calibration point -1mm/+2mm (green arrows). Thorlabs base PY005/M, reproduced with permission. The numbers are the sensor numbers i.e. SENSOR1 = 1. \textbf{(b)} Angle variations from 0° to 0.498° (rounded to 0.5°) in steps of 0.166°. \textbf{(c)} The hardware setup.}
  \label{fig:sensor-setup} 
\end{figure*}

The sensor array was first validated statically to understand its capabilities. A full factorial design of experiments was performed with three samples made of Si substrate and a layer of SiO\textsubscript{2} coating (Si:SiO\textsubscript{2}). Each sample had the following coating thicknesses: SAMPLE1: 300nm, SAMPLE2: 286nm and SAMPLE3: 164nm. The set of experiments consisted of calibrating at 2mm above the sample surface, then modifying the array height: -1mm/+2mm, and the wafer angle up to 0.498° (rounded to 0.5°) in increments of 0.166° as shown in Figure \ref{fig:sensor-setup}. 

\begin{table}[t]
    \centering
    \begin{adjustbox}{width=\textwidth, width = 0.7\textheight}
		\begin{tabular}{ccc|cccccccccccc|}
			\hline
			\multicolumn{3}{|c|}{} & \multicolumn{12}{c|}{\cellcolor[HTML]{DDEBF7}ARRAY HEIGHT (mm)}                                                                                                                                                                                                                \\ \cline{4-15} 
			\multicolumn{3}{|c|}{\multirow{-2}{*}{Position}}                                                                                                         & \multicolumn{1}{c|}{\cellcolor[HTML]{DDEBF7}1}      & \multicolumn{1}{c|}{\cellcolor[HTML]{DDEBF7}2*}     & \multicolumn{1}{c|}{\cellcolor[HTML]{DDEBF7}3}      & \multicolumn{1}{c|}{\cellcolor[HTML]{DDEBF7}4}      & \multicolumn{1}{c|}{\cellcolor[HTML]{DDEBF7}1}      & \multicolumn{1}{c|}{\cellcolor[HTML]{DDEBF7}2*}     & \multicolumn{1}{c|}{\cellcolor[HTML]{DDEBF7}3}      & \multicolumn{1}{c|}{\cellcolor[HTML]{DDEBF7}4}      & \multicolumn{1}{c|}{\cellcolor[HTML]{DDEBF7}1}      & \multicolumn{1}{c|}{\cellcolor[HTML]{DDEBF7}2*}     & \multicolumn{1}{c|}{\cellcolor[HTML]{DDEBF7}3}      & \cellcolor[HTML]{DDEBF7}4      \\ \hline
			\multicolumn{1}{|c|}{\cellcolor[HTML]{BFBFBF}}                        & \multicolumn{1}{c|}{}                              & \cellcolor[HTML]{EDEDED}0.5  & \multicolumn{1}{c|}{\cellcolor[HTML]{FFF2CC}0.0316} & \multicolumn{1}{c|}{\cellcolor[HTML]{FFF2CC}0.0241} & \multicolumn{1}{c|}{\cellcolor[HTML]{E2EFD9}0.0194} & \multicolumn{1}{c|}{\cellcolor[HTML]{E2EFD9}0.0176} & \multicolumn{1}{c|}{\cellcolor[HTML]{FFF2CC}0.6693} & \multicolumn{1}{c|}{\cellcolor[HTML]{FFF2CC}0.8665} & \multicolumn{1}{c|}{\cellcolor[HTML]{E2EFD9}0.9312} & \multicolumn{1}{c|}{\cellcolor[HTML]{E2EFD9}0.9416} & \multicolumn{1}{c|}{\cellcolor[HTML]{FFF2CC}304.55} & \multicolumn{1}{c|}{\cellcolor[HTML]{FFF2CC}301.13} & \multicolumn{1}{c|}{\cellcolor[HTML]{E2EFD9}300.6}  & \cellcolor[HTML]{E2EFD9}300.76                     \\ \cline{3-15} 
			\multicolumn{1}{|c|}{\cellcolor[HTML]{BFBFBF}}                        & \multicolumn{1}{c|}{}                              & \cellcolor[HTML]{EDEDED}0.33 & \multicolumn{1}{c|}{\cellcolor[HTML]{FFF2CC}0.0258} & \multicolumn{1}{c|}{\cellcolor[HTML]{E2EFD9}0.0188} & \multicolumn{1}{c|}{\cellcolor[HTML]{E2EFD9}0.0157} & \multicolumn{1}{c|}{\cellcolor[HTML]{E2EFD9}0.0171} & \multicolumn{1}{c|}{\cellcolor[HTML]{FFF2CC}0.8202} & \multicolumn{1}{c|}{\cellcolor[HTML]{E2EFD9}0.9315} & \multicolumn{1}{c|}{\cellcolor[HTML]{E2EFD9}0.9571} & \multicolumn{1}{c|}{\cellcolor[HTML]{E2EFD9}0.9433} & \multicolumn{1}{c|}{\cellcolor[HTML]{FFF2CC}301.46} & \multicolumn{1}{c|}{\cellcolor[HTML]{E2EFD9}301.58} & \multicolumn{1}{c|}{\cellcolor[HTML]{E2EFD9}301.11} & \cellcolor[HTML]{E2EFD9}300.47                     \\ \cline{3-15} 
			\multicolumn{1}{|c|}{\multirow{-3}{*}{\cellcolor[HTML]{BFBFBF}RIGHT}} & \multicolumn{1}{c|}{}                              & \cellcolor[HTML]{EDEDED}0.16 & \multicolumn{1}{c|}{\cellcolor[HTML]{E2EFD9}0.0218} & \multicolumn{1}{c|}{\cellcolor[HTML]{E2EFD9}0.0141} & \multicolumn{1}{c|}{\cellcolor[HTML]{E2EFD9}0.0142} & \multicolumn{1}{c|}{\cellcolor[HTML]{E2EFD9}0.018}  & \multicolumn{1}{c|}{\cellcolor[HTML]{E2EFD9}0.9086} & \multicolumn{1}{c|}{\cellcolor[HTML]{E2EFD9}0.9707} & \multicolumn{1}{c|}{\cellcolor[HTML]{E2EFD9}0.9643} & \multicolumn{1}{c|}{\cellcolor[HTML]{E2EFD9}0.9408} & \multicolumn{1}{c|}{\cellcolor[HTML]{E2EFD9}300.65} & \multicolumn{1}{c|}{\cellcolor[HTML]{E2EFD9}300.93} & \multicolumn{1}{c|}{\cellcolor[HTML]{E2EFD9}300.84} & \cellcolor[HTML]{E2EFD9}300.07                     \\ \cline{1-1} \cline{3-15} 
			\multicolumn{1}{|c|}{\cellcolor[HTML]{BFBFBF}CENTER}                  & \multicolumn{1}{c|}{}                              & \cellcolor[HTML]{EDEDED}0    & \multicolumn{1}{c|}{\cellcolor[HTML]{E2EFD9}0.0207} & \multicolumn{1}{c|}{\cellcolor[HTML]{E2EFD9}0.0124} & \multicolumn{1}{c|}{\cellcolor[HTML]{E2EFD9}0.0146} & \multicolumn{1}{c|}{\cellcolor[HTML]{E2EFD9}0.0197} & \multicolumn{1}{c|}{\cellcolor[HTML]{E2EFD9}0.927}  & \multicolumn{1}{c|}{\cellcolor[HTML]{E2EFD9}0.9791} & \multicolumn{1}{c|}{\cellcolor[HTML]{E2EFD9}0.9644} & \multicolumn{1}{c|}{\cellcolor[HTML]{E2EFD9}0.9305} & \multicolumn{1}{c|}{\cellcolor[HTML]{E2EFD9}299.69} & \multicolumn{1}{c|}{\cellcolor[HTML]{E2EFD9}300.08} & \multicolumn{1}{c|}{\cellcolor[HTML]{E2EFD9}299.34} & \cellcolor[HTML]{E2EFD9}300.12                     \\ \cline{1-1} \cline{3-15} 
			\multicolumn{1}{|c|}{\cellcolor[HTML]{BFBFBF}}                        & \multicolumn{1}{c|}{}                              & \cellcolor[HTML]{EDEDED}0.16 & \multicolumn{1}{c|}{\cellcolor[HTML]{E2EFD9}0.0204} & \multicolumn{1}{c|}{\cellcolor[HTML]{E2EFD9}0.0135} & \multicolumn{1}{c|}{\cellcolor[HTML]{E2EFD9}0.0158} & \multicolumn{1}{c|}{\cellcolor[HTML]{E2EFD9}0.0208} & \multicolumn{1}{c|}{\cellcolor[HTML]{E2EFD9}0.919}  & \multicolumn{1}{c|}{\cellcolor[HTML]{E2EFD9}0.9732} & \multicolumn{1}{c|}{\cellcolor[HTML]{E2EFD9}0.9571} & \multicolumn{1}{c|}{\cellcolor[HTML]{E2EFD9}0.9215} & \multicolumn{1}{c|}{\cellcolor[HTML]{E2EFD9}299.17} & \multicolumn{1}{c|}{\cellcolor[HTML]{E2EFD9}298.11} & \multicolumn{1}{c|}{\cellcolor[HTML]{E2EFD9}298.97} & \cellcolor[HTML]{E2EFD9}299.63                     \\ \cline{3-15} 
			\multicolumn{1}{|c|}{\cellcolor[HTML]{BFBFBF}}                        & \multicolumn{1}{c|}{}                              & \cellcolor[HTML]{EDEDED}0.33 & \multicolumn{1}{c|}{\cellcolor[HTML]{FFF2CC}0.022}  & \multicolumn{1}{c|}{\cellcolor[HTML]{E2EFD9}0.0171} & \multicolumn{1}{c|}{\cellcolor[HTML]{E2EFD9}0.0179} & \multicolumn{1}{c|}{\cellcolor[HTML]{E2EFD9}0.0218} & \multicolumn{1}{c|}{\cellcolor[HTML]{FFF2CC}0.8739} & \multicolumn{1}{c|}{\cellcolor[HTML]{E2EFD9}0.9486} & \multicolumn{1}{c|}{\cellcolor[HTML]{E2EFD9}0.943}  & \multicolumn{1}{c|}{\cellcolor[HTML]{E2EFD9}0.9116} & \multicolumn{1}{c|}{\cellcolor[HTML]{FFF2CC}299.55} & \multicolumn{1}{c|}{\cellcolor[HTML]{E2EFD9}297.82} & \multicolumn{1}{c|}{\cellcolor[HTML]{E2EFD9}298.94} & \cellcolor[HTML]{E2EFD9}298.64                     \\ \cline{3-15} 
			\multicolumn{1}{|c|}{\multirow{-3}{*}{\cellcolor[HTML]{BFBFBF}LEFT}}  & \multicolumn{1}{c|}{}                              & \cellcolor[HTML]{EDEDED}0.5  & \multicolumn{1}{c|}{\cellcolor[HTML]{FFF2CC}0.026}  & \multicolumn{1}{c|}{\cellcolor[HTML]{E2EFD9}0.0212} & \multicolumn{1}{c|}{\cellcolor[HTML]{E2EFD9}0.0208} & \multicolumn{1}{c|}{\cellcolor[HTML]{E2EFD9}0.0225} & \multicolumn{1}{c|}{\cellcolor[HTML]{FFF2CC}0.788}  & \multicolumn{1}{c|}{\cellcolor[HTML]{E2EFD9}0.9036} & \multicolumn{1}{c|}{\cellcolor[HTML]{E2EFD9}0.9209} & \multicolumn{1}{c|}{\cellcolor[HTML]{E2EFD9}0.9018} & \multicolumn{1}{c|}{\cellcolor[HTML]{FFF2CC}298.1}  & \multicolumn{1}{c|}{\cellcolor[HTML]{E2EFD9}298.37} & \multicolumn{1}{c|}{\cellcolor[HTML]{E2EFD9}298.77} & \cellcolor[HTML]{E2EFD9}298.45                     \\ \cline{1-1} \cline{3-15} 
			\multicolumn{1}{|c|}{\cellcolor[HTML]{BFBFBF}}                        & \multicolumn{1}{c|}{}                              & \cellcolor[HTML]{EDEDED}0.5  & \multicolumn{1}{c|}{\cellcolor[HTML]{FFF2CC}0.0271} & \multicolumn{1}{c|}{\cellcolor[HTML]{E2EFD9}0.0217} & \multicolumn{1}{c|}{\cellcolor[HTML]{E2EFD9}0.0202} & \multicolumn{1}{c|}{\cellcolor[HTML]{E2EFD9}0.0203} & \multicolumn{1}{c|}{\cellcolor[HTML]{FFF2CC}0.7751} & \multicolumn{1}{c|}{\cellcolor[HTML]{E2EFD9}0.9051} & \multicolumn{1}{c|}{\cellcolor[HTML]{E2EFD9}0.9292} & \multicolumn{1}{c|}{\cellcolor[HTML]{E2EFD9}0.9284} & \multicolumn{1}{c|}{\cellcolor[HTML]{FFF2CC}299.56} & \multicolumn{1}{c|}{\cellcolor[HTML]{E2EFD9}300.34} & \multicolumn{1}{c|}{\cellcolor[HTML]{E2EFD9}301.36} & \cellcolor[HTML]{E2EFD9}301.12                     \\ \cline{3-15} 
			\multicolumn{1}{|c|}{\cellcolor[HTML]{BFBFBF}}                        & \multicolumn{1}{c|}{}                              & \cellcolor[HTML]{EDEDED}0.33 & \multicolumn{1}{c|}{\cellcolor[HTML]{FFF2CC}0.0236} & \multicolumn{1}{c|}{\cellcolor[HTML]{E2EFD9}0.018}  & \multicolumn{1}{c|}{\cellcolor[HTML]{E2EFD9}0.0173} & \multicolumn{1}{c|}{\cellcolor[HTML]{E2EFD9}0.0191} & \multicolumn{1}{c|}{\cellcolor[HTML]{FFF2CC}0.8575} & \multicolumn{1}{c|}{\cellcolor[HTML]{E2EFD9}0.9409} & \multicolumn{1}{c|}{\cellcolor[HTML]{E2EFD9}0.9486} & \multicolumn{1}{c|}{\cellcolor[HTML]{E2EFD9}0.9325} & \multicolumn{1}{c|}{\cellcolor[HTML]{FFF2CC}300.05} & \multicolumn{1}{c|}{\cellcolor[HTML]{E2EFD9}299.82} & \multicolumn{1}{c|}{\cellcolor[HTML]{E2EFD9}300.23} & \cellcolor[HTML]{E2EFD9}300.53                     \\ \cline{3-15} 
			\multicolumn{1}{|c|}{\multirow{-3}{*}{\cellcolor[HTML]{BFBFBF}BACK}}  & \multicolumn{1}{c|}{}                              & \cellcolor[HTML]{EDEDED}0.16 & \multicolumn{1}{c|}{\cellcolor[HTML]{E2EFD9}0.0207} & \multicolumn{1}{c|}{\cellcolor[HTML]{E2EFD9}0.014}  & \multicolumn{1}{c|}{\cellcolor[HTML]{E2EFD9}0.0154} & \multicolumn{1}{c|}{\cellcolor[HTML]{E2EFD9}0.0186} & \multicolumn{1}{c|}{\cellcolor[HTML]{E2EFD9}0.915}  & \multicolumn{1}{c|}{\cellcolor[HTML]{E2EFD9}0.9685} & \multicolumn{1}{c|}{\cellcolor[HTML]{E2EFD9}0.9589} & \multicolumn{1}{c|}{\cellcolor[HTML]{E2EFD9}0.9357} & \multicolumn{1}{c|}{\cellcolor[HTML]{E2EFD9}299.38} & \multicolumn{1}{c|}{\cellcolor[HTML]{E2EFD9}299.79} & \multicolumn{1}{c|}{\cellcolor[HTML]{E2EFD9}299.74} & \cellcolor[HTML]{E2EFD9}300.08                     \\ \cline{1-1} \cline{3-15} 
			\multicolumn{1}{|c|}{\cellcolor[HTML]{BFBFBF}CENTER}                  & \multicolumn{1}{c|}{}                              & \cellcolor[HTML]{EDEDED}0    & \multicolumn{1}{c|}{\cellcolor[HTML]{E2EFD9}0.0189} & \multicolumn{1}{c|}{\cellcolor[HTML]{E2EFD9}0.0115} & \multicolumn{1}{c|}{\cellcolor[HTML]{E2EFD9}0.0142} & \multicolumn{1}{c|}{\cellcolor[HTML]{E2EFD9}0.0187} & \multicolumn{1}{c|}{\cellcolor[HTML]{E2EFD9}0.9434} & \multicolumn{1}{c|}{\cellcolor[HTML]{E2EFD9}0.9805} & \multicolumn{1}{c|}{\cellcolor[HTML]{E2EFD9}0.965}  & \multicolumn{1}{c|}{\cellcolor[HTML]{E2EFD9}0.9368} & \multicolumn{1}{c|}{\cellcolor[HTML]{E2EFD9}298.79} & \multicolumn{1}{c|}{\cellcolor[HTML]{E2EFD9}299.32} & \multicolumn{1}{c|}{\cellcolor[HTML]{E2EFD9}299.91} & \cellcolor[HTML]{E2EFD9}300.76                     \\ \cline{1-1} \cline{3-15} 
			\multicolumn{1}{|c|}{\cellcolor[HTML]{BFBFBF}}                        & \multicolumn{1}{c|}{}                              & \cellcolor[HTML]{EDEDED}0.16 & \multicolumn{1}{c|}{\cellcolor[HTML]{FFF2CC}0.0242} & \multicolumn{1}{c|}{\cellcolor[HTML]{E2EFD9}0.0127} & \multicolumn{1}{c|}{\cellcolor[HTML]{E2EFD9}0.0132} & \multicolumn{1}{c|}{\cellcolor[HTML]{E2EFD9}0.0181} & \multicolumn{1}{c|}{\cellcolor[HTML]{FFF2CC}0.8937} & \multicolumn{1}{c|}{\cellcolor[HTML]{E2EFD9}0.9776} & \multicolumn{1}{c|}{\cellcolor[HTML]{E2EFD9}0.9701} & \multicolumn{1}{c|}{\cellcolor[HTML]{E2EFD9}0.94}   & \multicolumn{1}{c|}{\cellcolor[HTML]{E2EFD9}299.01} & \multicolumn{1}{c|}{\cellcolor[HTML]{E2EFD9}299.18} & \multicolumn{1}{c|}{\cellcolor[HTML]{E2EFD9}299.05} & \cellcolor[HTML]{E2EFD9}300.26                     \\ \cline{3-15} 
			\multicolumn{1}{|c|}{\cellcolor[HTML]{BFBFBF}}                        & \multicolumn{1}{c|}{}                              & \cellcolor[HTML]{EDEDED}0.33 & \multicolumn{1}{c|}{\cellcolor[HTML]{FFF2CC}0.0267} & \multicolumn{1}{c|}{\cellcolor[HTML]{E2EFD9}0.015}  & \multicolumn{1}{c|}{\cellcolor[HTML]{E2EFD9}0.0135} & \multicolumn{1}{c|}{\cellcolor[HTML]{E2EFD9}0.0174} & \multicolumn{1}{c|}{\cellcolor[HTML]{FFF2CC}0.8504} & \multicolumn{1}{c|}{\cellcolor[HTML]{E2EFD9}0.9664} & \multicolumn{1}{c|}{\cellcolor[HTML]{E2EFD9}0.9686} & \multicolumn{1}{c|}{\cellcolor[HTML]{E2EFD9}0.944}  & \multicolumn{1}{c|}{\cellcolor[HTML]{FFF2CC}299.09} & \multicolumn{1}{c|}{\cellcolor[HTML]{E2EFD9}298.83} & \multicolumn{1}{c|}{\cellcolor[HTML]{E2EFD9}299.29} & \cellcolor[HTML]{E2EFD9}300.48                     \\ \cline{3-15} 
			\multicolumn{1}{|c|}{\multirow{-3}{*}{\cellcolor[HTML]{BFBFBF}FRONT}} & \multicolumn{1}{c|}{\multirow{-14}{*}{ANGLE (°)}} & \cellcolor[HTML]{EDEDED}0.5  & \multicolumn{1}{c|}{\cellcolor[HTML]{FFF2CC}0.0341} & \multicolumn{1}{c|}{\cellcolor[HTML]{E2EFD9}0.0202} & \multicolumn{1}{c|}{\cellcolor[HTML]{E2EFD9}0.0157} & \multicolumn{1}{c|}{\cellcolor[HTML]{E2EFD9}0.0166} & \multicolumn{1}{c|}{\cellcolor[HTML]{FFF2CC}0.7132} & \multicolumn{1}{c|}{\cellcolor[HTML]{E2EFD9}0.9311} & \multicolumn{1}{c|}{\cellcolor[HTML]{E2EFD9}0.9616} & \multicolumn{1}{c|}{\cellcolor[HTML]{E2EFD9}0.9481} & \multicolumn{1}{c|}{\cellcolor[HTML]{FFF2CC}301.33} & \multicolumn{1}{c|}{\cellcolor[HTML]{E2EFD9}298.39} & \multicolumn{1}{c|}{\cellcolor[HTML]{E2EFD9}299.24} & \cellcolor[HTML]{E2EFD9}299.56                     \\ \hline
			\multicolumn{1}{c|}{}                                                 & \multicolumn{2}{c|}{}                                                            & \multicolumn{4}{c|}{RMSE}                                                                                                                                                                                             & \multicolumn{4}{c|}{$R^2$}                                                                                                                                                                                            & \multicolumn{4}{c|}{Thickness (nm)}                                                                                                                                                              \\ \cline{2-15} 
		\end{tabular}
	\end{adjustbox}
    \caption{Full factorial DOE of SAMPLE1 (300nm) showing the averaged RMSE, $R^2$ and Thickness (nm) per combination of factors (Angle vs Height). Notes: (i) The Thorlabs PY005/M base was re-positioned when measuring RIGHT-LEFT and FRONT-BACK positions. (ii) Each data value is an average of thirty readings performed by all the sensors. (iii) *Calibration point @ height = 2mm from the sample surface.}
    \label{tab:sample-1}
\end{table}

Table \ref{tab:sample-1} shows the SAMPLE1 sensor-array measurements with increments of 0.166$^{\circ}$ rounded to the nearest 2 decimal places: RMSE, $R^2$ and Thickness. It was observed that when the RMSE was $\leq$ 0.022, the $R^2 > 0.9$. This is considered a good result for the sensor array as it shows that the measured reflectance curve for each sensor fits correctly to the modelled reflectance curve, as explained in the methods section. By contrast, when the RMSE is > 0.022, the $R^2$ likely drops below or trends towards 0.9. When this occurs some of the individual sensors present a loss of performance.

The loss of performance was first observed when the sensor array was positioned at the calibration point and when there was a 0.5° tilt on the right side of the sensor array. After reviewing the individual sensor performance it is clear that when the array tilts to the right side, the SENSOR6 showed an RMSE > 0.04 and $R^2$ < 0.7. Then, when varying the height 1mm below the calibration point and tilting the sensor array to the right side by 0.33°, SENSOR5 and SENSOR6 showed the same behaviour. Similarly, when there is a tilt on the left side below the calibration point, SENSOR1 and SENSOR7 showed an increase in RMSE and a decrease in $R^2$ when varying the angle 0.33°. Comparably, SENSOR6 and SENSOR7 detected variation when the wafer was tilted on the back side of the sensor array. Finally, all sensors except SENSOR2 detected a variation on the front side of the sensor array. This behaviour was repeatable for all the samples, data for which can be found in Supplementary 1.

Reduced performance of the RMSE and $R^2$ per sensor is caused by a change in the reflected intensity received by the sensor slit due to the angle and height variations distorting the measured reflectance curve. When this occurs, the RMSE increases above 0.4, similarly, the $R^2$ calculation gets affected because the optimisation algorithm fails to effectively fit a line within the measured reflectance curve making the $R^2$ go below <0.7. See the Methods section for more details on $R^2$ calculation.

Finally, despite the loss of the fit quality metrics, the final observation was that all combinations show an estimated thickness of less than 2\% variation (<6nm) vs the expected thickness value of 300nm. However, when the RMSE and $R^2$ fail beyond the expected levels, the thickness values are not reliable and sensor alignment to the calibration point must be performed.

\begin{figure}[h]
    \centering
    \includegraphics[width=0.5\textwidth]{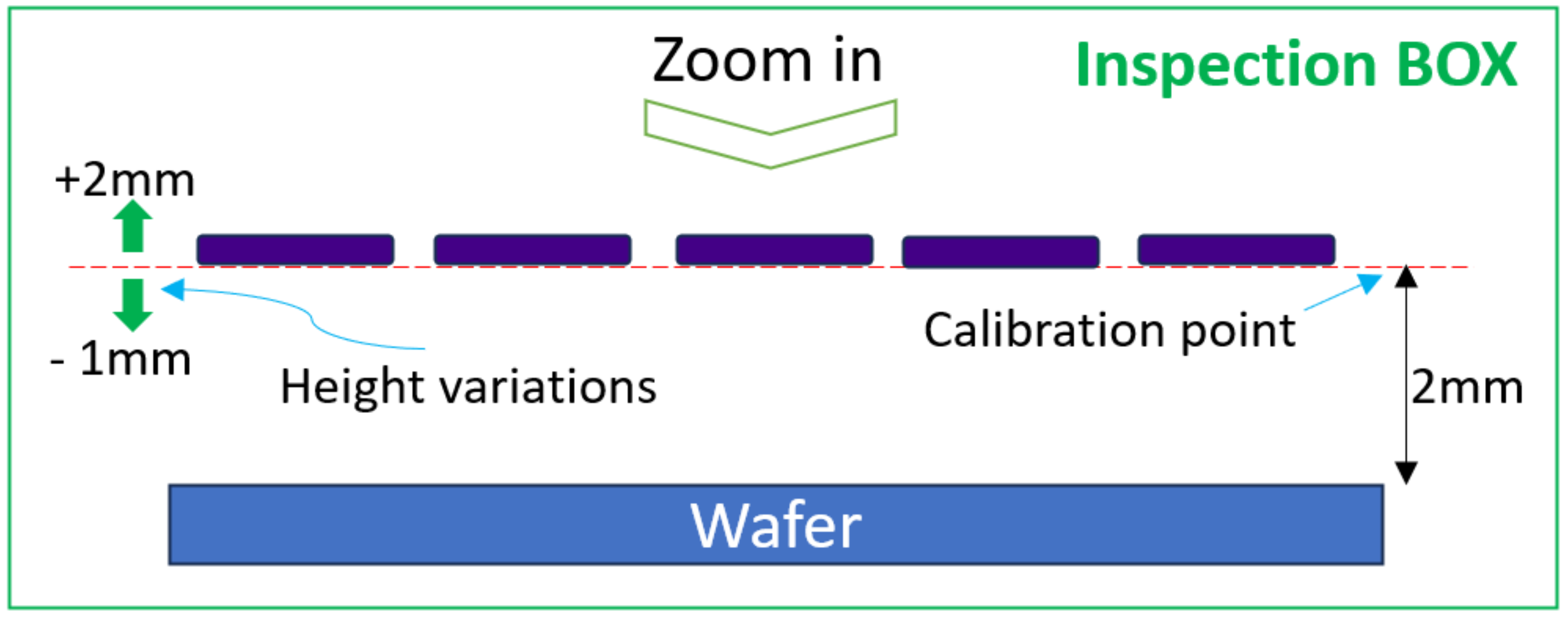}
    \caption{Sensor array inspection box}
    \label{fig:calib-box}
\end{figure}

Based on the static experiment results, an inspection box for the presented sensor array was defined as follows: The virtual inspection box is the sensor array limit to measure thin film thickness, whereas the inspection box is the baseline for an automated inspection procedure using a robot manipulator. A model example of the inspection box is found in \ref{fig:calib-box}.

\subsubsection*{Sample Comparison with Filmetrics F20}

An additional test was performed to evaluate the capability of the sensor array to meet manufacturing tolerances. The performance of the array was compared to a Filmetrics F20 reflectometer using three samples, with results shown in table \ref{tab:array-compairson}

\begin{table}[h]
\centering
\begin{tabular}{|c|c|c|c|c|c|c|}
\hline
Sample & \begin{tabular}[c]{@{}c@{}}Wafer\\ Vendor\end{tabular} & \begin{tabular}[c]{@{}c@{}}Thickness \\ (mm)\end{tabular} & \begin{tabular}[c]{@{}c@{}}Vendor\\ Tolerance (\%)\end{tabular} & \begin{tabular}[c]{@{}c@{}}Measured Thickness \\ - Array (nm)\end{tabular} & \begin{tabular}[c]{@{}c@{}}Measured Thickness\\  - Filmetrics F20 (nm)\end{tabular} & \begin{tabular}[c]{@{}c@{}}Array Error \\  vs Filmetrics (\%)\end{tabular} \\ \hline
1 & Pi-Kem & 300 & $\pm$ 20 & 300 & 303.9 & 1.3 \\ 
\hline
2 & Pi-Kem & 300 & 20 & 286 & 285.79 & 0.07 \\ 
\hline
3 & Inseto & 150 & 10 & 165 & 161.36 & 1.64 \\ 
\hline
\end{tabular}
\caption{\label{tab:array-compairson} Comparison of the proposed sensor array against the Filmetrics F20 reflectometer.}
\end{table}

The sensor array measurements demonstrated that the samples meet the vendor's tolerances. Two hundred and forty measurements were performed per sample in the same area positions as the sensor array and the F20 showed similar thickness measurements demonstrating that the sensor array is capable of measuring thickness values with an error below 2\% compared to the F20 reflectometer, which demonstrates the capability of our developed sensing array with respect to the reflectometer. 

\subsubsection*{Dynamic Measurement Accuracy}

\begin{figure*}[t]
    \centering
  \subfloat[\label{fig:motion}]{%
       \includegraphics[width=0.8\linewidth]{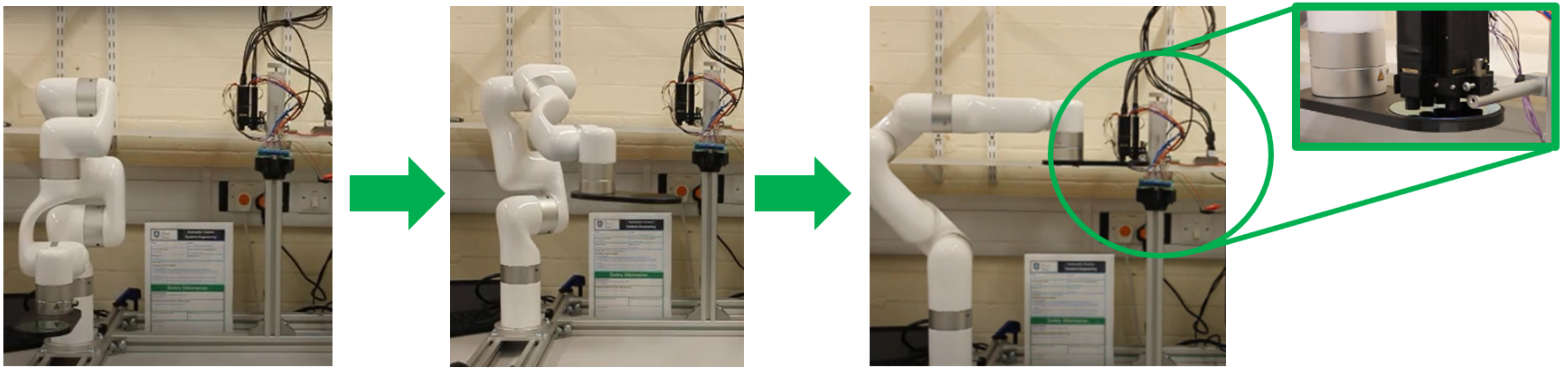}} \\
  \subfloat[\label{fig:robot-box}]{%
        \includegraphics[width=0.25\linewidth]{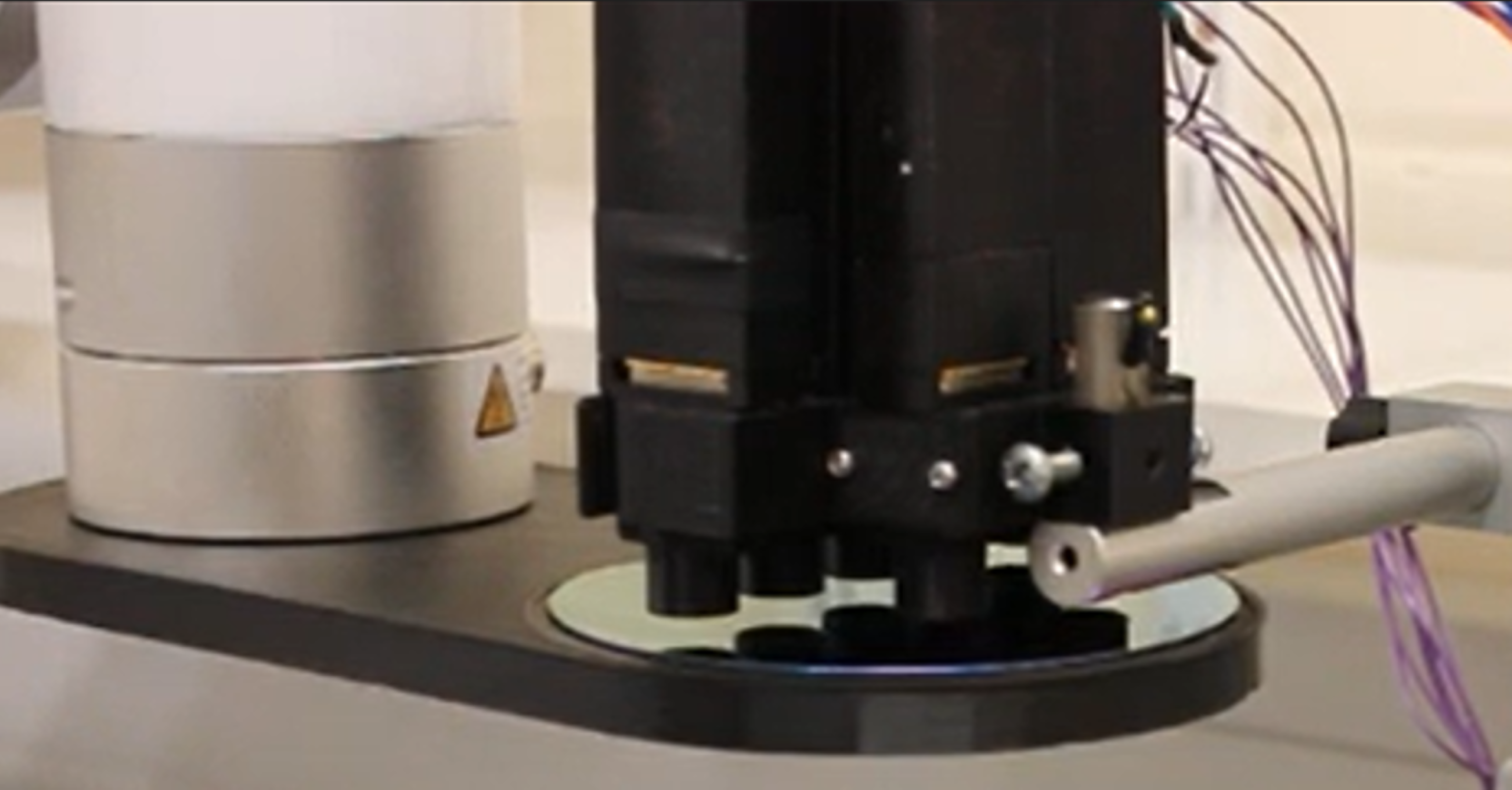}}
    \hspace{0.5cm}
  \subfloat[\label{fig:circular}]{%
        \includegraphics[width=0.25\linewidth]{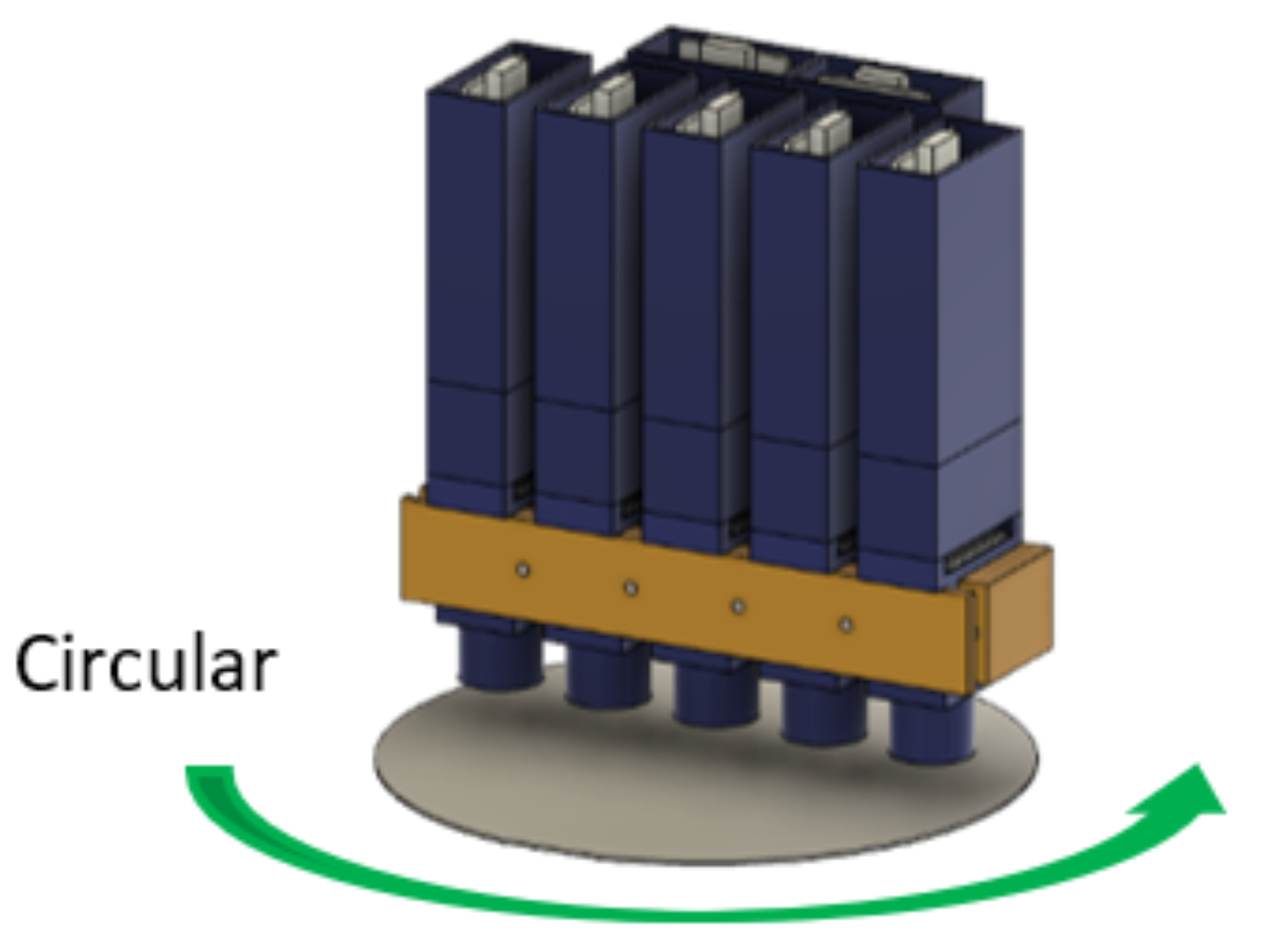}}
    \hspace{0.5cm}
  \subfloat[\label{fig:back-forth}]{%
        \includegraphics[width=0.25\linewidth]{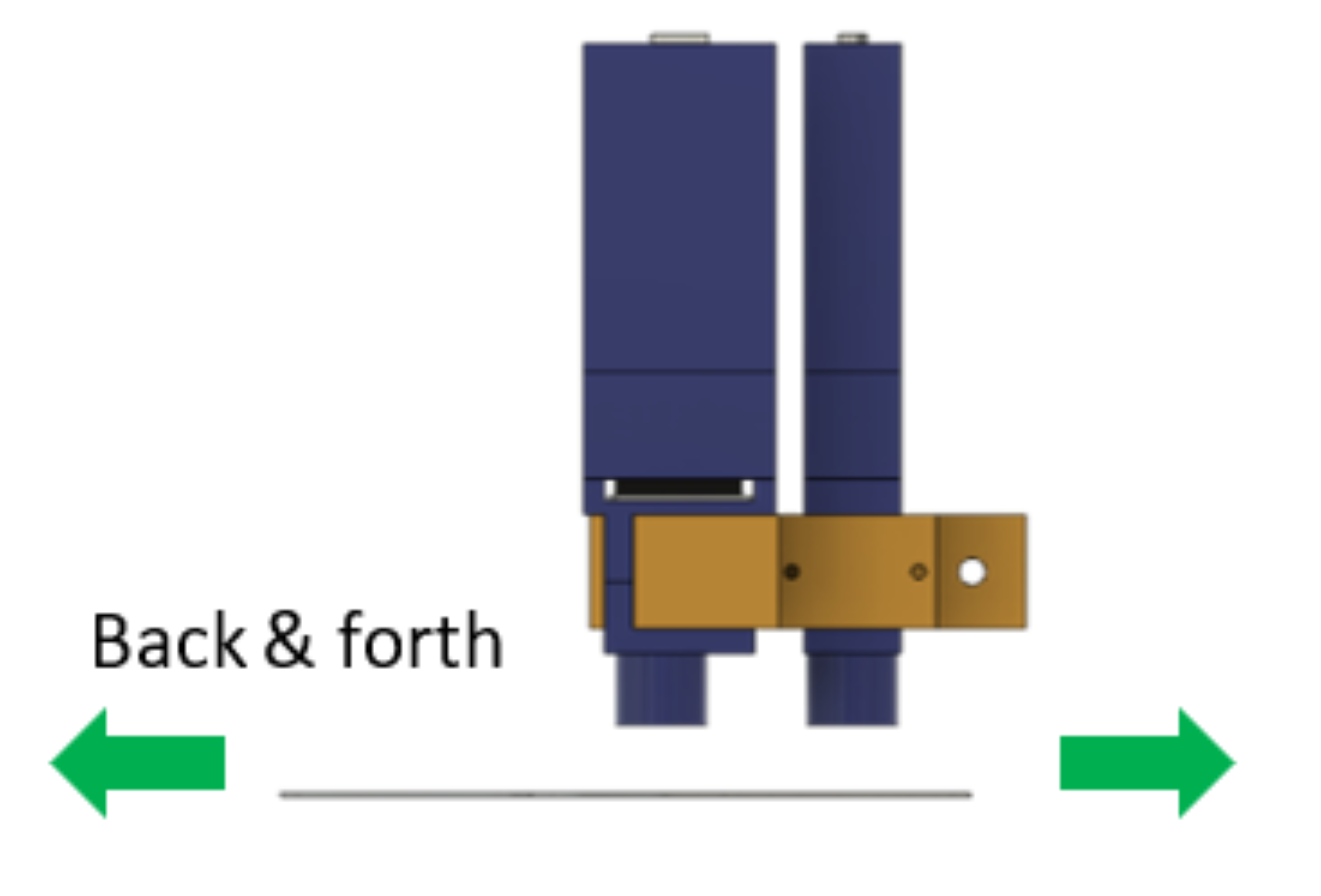}}
  \caption{Wafer inspection experiments with robot manipulator: \textbf{(a)} Robot arm movement from an idle/load position to the calibration point. The green rectangle shows the wafer and a 3D-printed end effector. \textbf{(b)} Zoom-in of the end effector and the wafer. \textbf{(c)} Circular trajectory. \textbf{(d)} Back \& forth trajectory}
  \label{fig:robot-sensing} 
\end{figure*}

Four automated trajectories were implemented with a robot manipulator, shown in figure \ref{fig:robot-sensing}. Figure \ref{fig:motion} shows the robot arm trajectory to the calibration position. First, the Si uncoated wafer is placed in the end effector and the arm moves it to the calibration point to start the calibration process. After taking the uncoated intensity, the arm goes back to the idle/load position to remove the Si wafer and then load the Si:SiO\textsubscript{2} coated wafer and complete the calibration process. Figure \ref{fig:robot-box} shows the wafer loaded on the end effector. Figure \ref{fig:circular} shows a circular motion to perform an area scan of the surface and figure \ref{fig:back-forth} shows the back-and-forth trajectory to simulate an R2R motion. The final trajectories are the positions shown in figure \ref{fig:setup-2}, used to duplicate the static experiment results and evaluate the inspection box calculations.

\begin{figure*}[t]
    \centering
  \subfloat[\label{fig:results-sample1}]{%
       \includegraphics[width=0.85\linewidth]{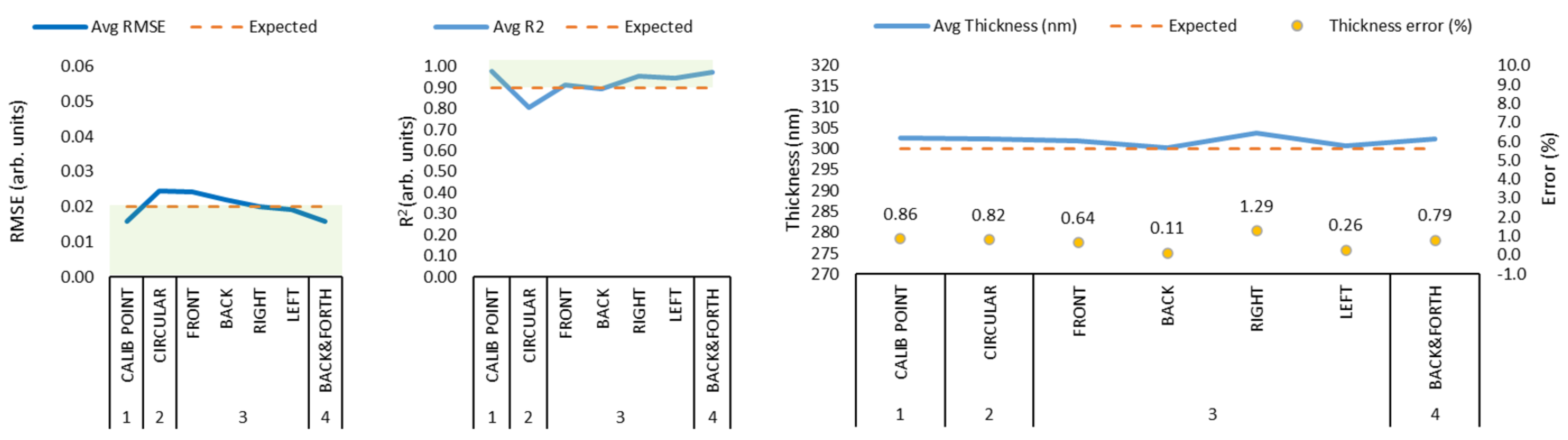}} \\
  \subfloat[\label{fig:results-sample2}]{%
        \includegraphics[width=0.85\linewidth]{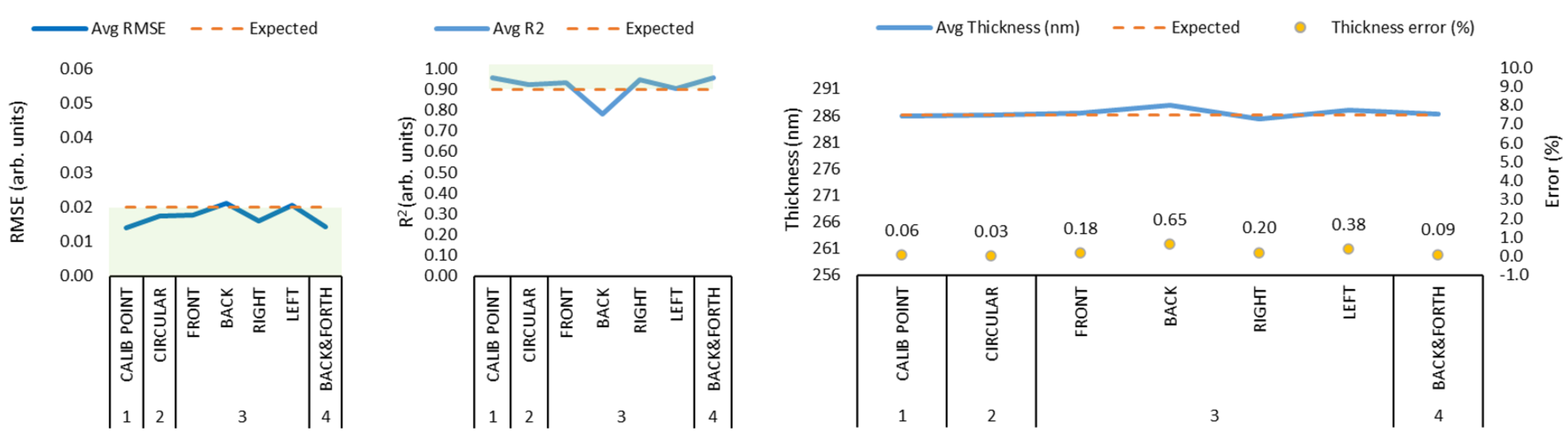}} \\ 
  \subfloat[\label{fig:results-sample3}]{%
        \includegraphics[width=0.85\linewidth]{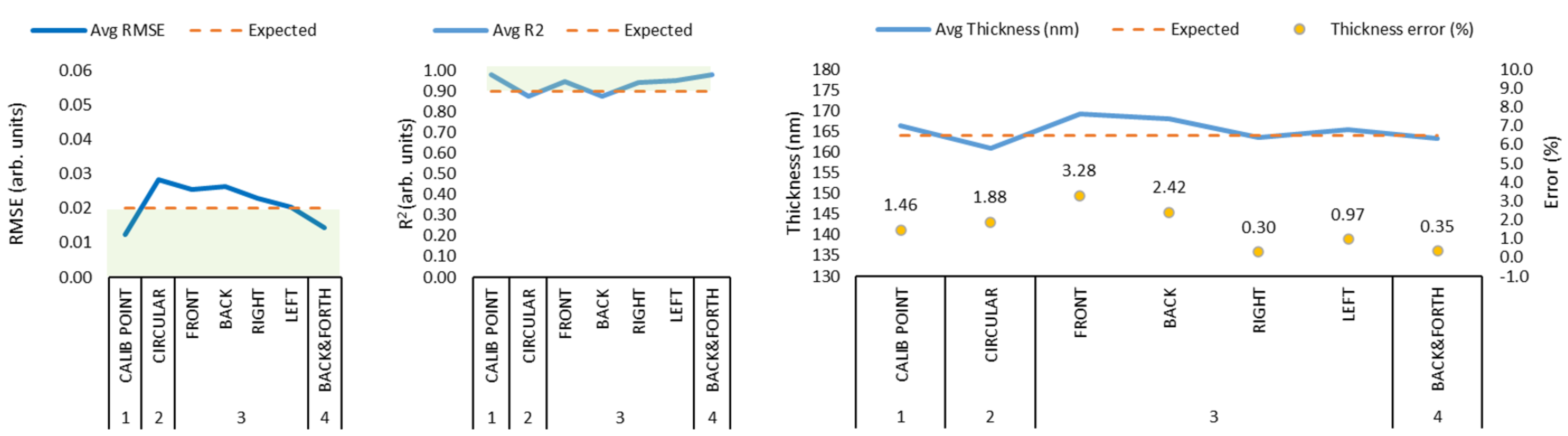}} \\
  \caption{Sensor array results (From left to right: RMSE, $R^2$ and Thickness). \textbf{(a)} SAMPLE1: 300nm, \textbf{(b)} SAMPLE2: 286nm, \textbf{(c)} SAMPLE3: 164nm. Notes: (i) The green zone in the RMSE and $R^2$ is the discovered safe zone for the array output. (ii) Each data point represents the averaged value of 50 measurements of all the sensors.}
  \label{fig:dynamic-results} 
\end{figure*}

Figure \ref{fig:dynamic-results} shows the sensor array results for each one of the programmed trajectories per sample. Figure \ref{fig:results-sample1} shows the SAMPLE1 (300nm) results. As observed in the figure, from left to right, Sequence 1 (CALIB POINT) shows an RMSE < 0.02, $R^2$ > 0.9 and thickness measurement = 302.57nm. This represents a thickness error of 0.86\% vs the expected thickness value of 300nm, which is close to the 0.4\% accuracy offered by the F20 and within the +/- 2\% compared to Yersak,etal for a potential R2R application \cite{yersak_2014}. This behaviour was similar when performing Sequence 4 (BACK AND FORTH) as it showed an error of 0.79\%. Nevertheless, Sequence 2 (CIRCULAR) showed an increase in the RMSE > 0.02, meaning that the sensor array detected height variations below the calibration point whilst performing the sequence. In this case, it was observed that SENSOR1, SENSOR2, and SENSOR6 started failing showing an RMSE > 0.04 and an $R^2$ < 0.07 after half of the circular motion sequence (Mov7) which suggests a misalignment of the end effector. When the RMSE and $R^2$ go beyond these limits, it was observed that the individual thickness measurements can go above 10\% of the expected thickness value. For instance, when SENSOR6 showed an RMSE=0.07 and $R^2$=0.63, the individual thickness reading was 368.26nm affecting the general average metric score. Sequence 3 was designed to duplicate the static experiments, for a potential in-motion tilt detection. The array could detect front and back tilting, with SENSOR5 and SENSOR7 detecting RMSE variations of >0.02 and $R^2$ close to 0.9, however, it was not possible to detect significant variations beyond the set limits when the robot arm tilted the wafer on the right and left sides. Similar behaviour was observed in SAMPLE2 (286nm) and SAMPLE3 (164nm) when performing all the robot arm programmed sequences, data for which can be found in Supplementary 2.

\subsection*{Verification of Robot Movement}

\subsubsection*{Constraint Manifold Identification}

\begin{figure*}[t]
    \centering
    \hfill
  \subfloat[\label{fig:elbo}]{%
       \includegraphics[width=0.3\linewidth]{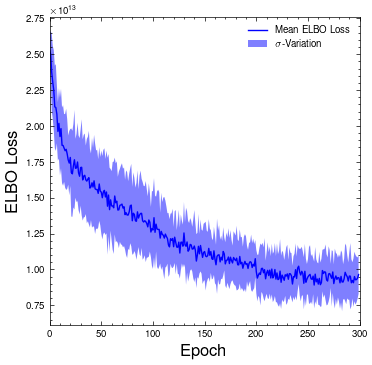}}
       \hfill
  \subfloat[\label{fig:var-measure}]{%
        \includegraphics[width=0.33\linewidth]{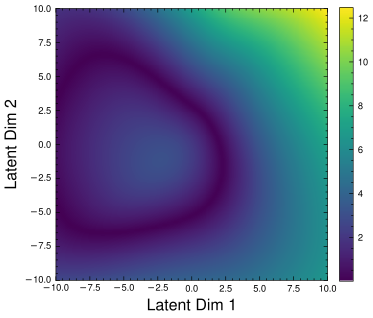}} 
        \hfill
  \subfloat[\label{fig:mag-fac}]{%
        \includegraphics[width=0.33\linewidth]{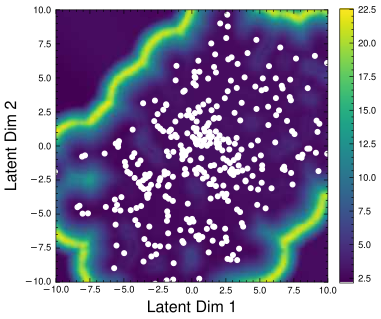}} 
  \caption{Training results from the Riemannian manifold VAE: \textbf{(a)} ELBO loss averaged across 10 runs, indicating convergence on a stable manifold; \textbf{(b)} Variance measure of the latent space. This variance takes low values in areas that the manifold has a high confidence of performance and high values in areas of high uncertainty; \textbf{(c)} The magnification factor $J$ in equation \ref{eq:mag-fac} applied to the variance measure metric. The white dots indicate the training data, with a boundary around those points of high variance indicating the edge of the manifold.}
  \label{fig:vae-results} 
\end{figure*}

To identify the constraint manifold, the first manifold that was obtained was the global manipulation manifold. The results of the training process are shown in figure \ref{fig:elbo}, where the variance measure in figure \ref{fig:var-measure} shows the confidence of the variational autoencoder (VAE) in its predictions of the joint space positions from the latent space into the joint space. In figure \ref{fig:mag-fac}, the magnification factor of the metric is defined as:
\begin{equation}\label{eq:mag-fac}
    J = \log \sqrt{\det [ \mathbf{M}_{J_\mu} + \mathbf{M}_{J_\sigma}]}
\end{equation}
\noindent where $\mathbf{M_{J_\mu}}$ and $\mathbf{M_{J_\sigma}}$ are the Jacobian matrices of the VAE's decoder mean and variance architecture outputs respectively, governing the model's confidence in it's estimation of the manifold. This metric shows the boundary of the manifold based on the data \cite{mag_fac_1997}, indicating that the lower dimensions constitute a manifold $\mathcal{M}_\theta$ encompassing the joint positions. It is observed that the latent space representation in figure \ref{fig:var-measure} correlates to the data spread presented in figure \ref{fig:mag-fac}'s distribution of embedded points on the manifold, indicating that the lower-dimensional representation is sufficient for the application of studying a manipulator's kinematics.

We can modify the learned manifold domain by applying the constraint function metric $\mathbf{M}^\theta_{\mathbf{f}}$ to the variance measure plot. This creates a sub-manifold $\tilde{\mathcal{M}}$ that satisfies the constraint function $\mathbf{f}(\hat{\theta}) = 0$, where $\hat{\theta}$ indicates a predicted value from the VAE. As our application is considering the manipulator to maintain horizontal motion during operation, we apply a constraint vector of $\mathbb{C} = [0\ 0\ 0\ 1\ 1\ 0]^\intercal$ then compute the value of $\mathbf{f}(\hat{\theta})$ for each point in the latent space. This is then used to determine the constraint metric $\mathbf{M}^\theta_{\mathbf{f}}$, which we can then project onto a three-dimensional variance measure plot.

\begin{figure}[h]
    \floatbox[{\capbeside\thisfloatsetup{capbesideposition={left,center},capbesidewidth=5cm}}]{figure}[\FBwidth]
    {\caption{Projected value of the constraint metric $\mathbf{M}^\theta_{\mathbf{f}}$, normalised to between 0 and 1, onto the variance measure in three-dimensions. Areas corresponding to zero (dark blue) are latent space points that satisfy the constraint. The color bar on the left represents the value of the constraint metric, with the $Z$-axis value being the variance metric.}\label{fig:metric-3d}}
    {\includegraphics[width=0.5\textwidth]{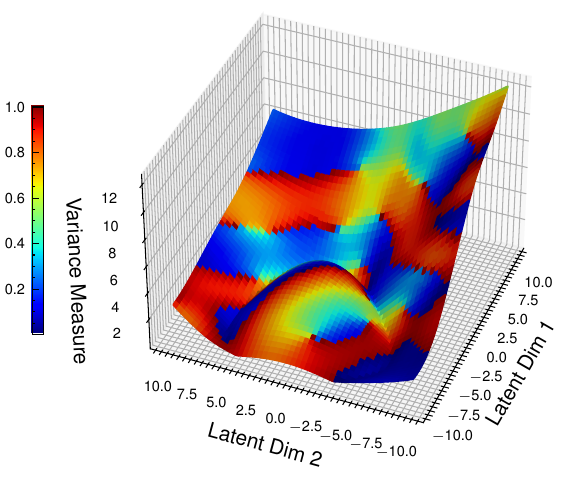}}
\end{figure}

As shown in figure \ref{fig:metric-3d}, there exists a sub-manifold - shown in dark blue - that corresponds to areas that satisfy the constraint vector and maintain horizontal motion. As these areas lie on the low variance regions of the manifold, we can use this learned model to examine the manipulator movement and determine with a high degree of confidence whether the constraints are being met during the motion of the manipulator.  

\subsubsection*{Movement Inspection}

To examine the ability of the latent space manifold to detect variations, we can embed motion plans generated on hardware into the latent space. This is done using the MoveIt motion planner in ROS \cite{moveit_2014}, where ground-truths are obtained for plans that are no longer adhering to constraints. These plans are then encoded into the latent space of the VAE to be evaluated with the constraint metric $\mathbf{M}^\theta_{\mathbf{f}}$. For each path generated in MoveIt, points were sampled from the path and labelled with the ground truth as to whether they satisfy the constraints. The paths were generated from the starting position to the calibration box shown in figures \ref{fig:setup-1} and \ref{fig:calib-box}, with three off-manifold paths being induced with increasing deviation from the desired ground-truth trajectory. 

\begin{figure}[t]
    \centering
    \includegraphics[width=0.65\textwidth]{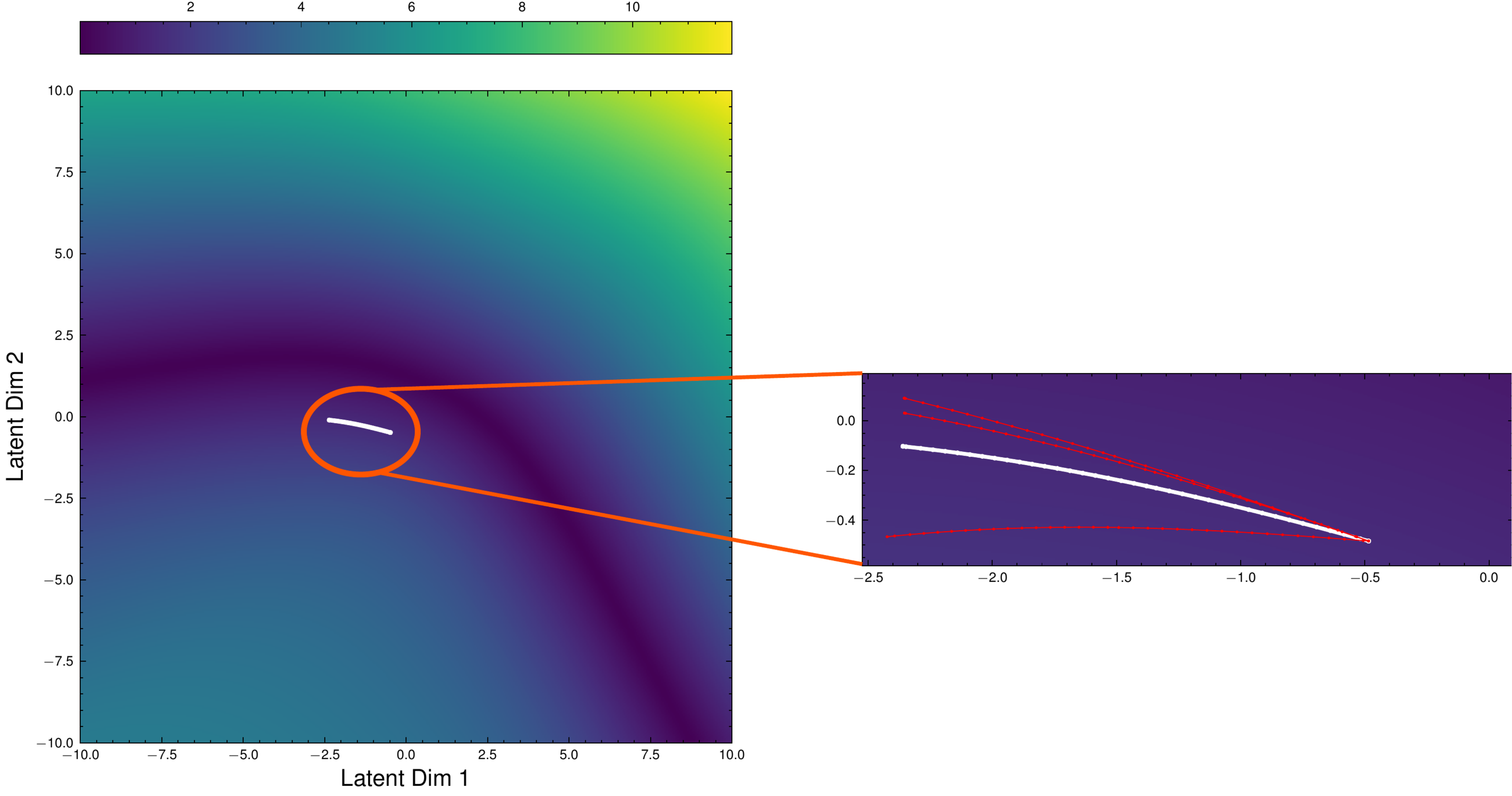}
    \caption{Projection of trajectories from the manipulator into the latent space of the VAE. \textbf{LEFT}: Projection of the trajectories on to the manifold; \textbf{RIGHT}: Zoomed in of the space where the trajectories lie on the manifold. The trajectory in white is the ground truth trajectory that maintains horizontal motion during the trajectory. Trajectories in red are ones that deviate from the constraint sub-manifold.}
    \label{fig:traj-proj}
\end{figure}

As shown in figure \ref{fig:traj-proj}, the trajectories can be shown in the latent space as being continuous, indicating that continuity with joint positions across single trajectories is maintained when embedding the high dimensional joint space onto the latent space Riemannian manifold. This continuity is caused by the fact that the VAE maintains the geometric relationships between the joint positions in Euclidean space when embedding them in the latent space. Furthermore, this preservation of the robot geometry is present when examining the trajectories that violate the constraint, shown in red in figure \ref{fig:traj-proj}, as there is significant deviation from the desired trajectory in white which indicates that trajectories that violate the constraint imposed on the manipulator can be detected in the latent space. This deviation is determined by examining the value of the metric $\mathbf{M}^\theta_{\mathbf{f}}$, which allows us to build the curvature of the manifold based on the value of the constraint in the latent space. 

\begin{table}[h]
    \centering
    \begin{tabular}{|c|cc|cc|}
    \hline
    \multirow{2}{*}{\textbf{Trajectory}} & \multicolumn{2}{c|}{\textbf{Reconstruction Accuracy}} & \multicolumn{2}{c|}{\textbf{Manifold Constraint Estimation}} \\ \cline{2-5} 
                                         & \multicolumn{1}{c|}{$\mu$} & $\sigma$ & \multicolumn{1}{c|}{$\mu$} & $\sigma$ \\ \hline
    On-Manifold                          & \multicolumn{1}{c|}{-0.18}           & 0.24           & \multicolumn{1}{c|}{-0.014}              & 1.81              \\ \hline
    Off-Manifold 1                       & \multicolumn{1}{c|}{-0.08}           & 0.23           & \multicolumn{1}{c|}{-1.30}               & 2.68              \\ \hline
    Off-Manifold 2                       & \multicolumn{1}{c|}{-0.003}          & 0.30           & \multicolumn{1}{c|}{0.82}                & 1.49              \\ \hline
    Off-Manifold 3                       & \multicolumn{1}{c|}{-0.16}           & 0.30           & \multicolumn{1}{c|}{-0.11}               & 2.28              \\ \hline
    \end{tabular}
    \caption{\label{tab:vae-comp} Evaluation of the trajectory reconstruction and direct constraint estimation difference between the ground truth and the values from the latent space manifold. The mean value $\mu$ is the average difference between the actual ground truth and the reconstructed output from the VAE with the standard deviation $\sigma$ over the trajectory.}
\end{table}

Using this projection, we can determine the estimated values of the constraint function directly from this manifold without needing to explicitly calculate its value. We can also evaluate the accuracy of the reconstructed points from the VAE decoder to determine the ability of the VAE architecture to learn the lower-dimensional manifold. From table \ref{tab:vae-comp}, we can see the VAE can reproduce the encoded positions with a high degree of accuracy, indicating that the latent space in the VAE is an accurate representation of the joint configurations for the manipulator. Furthermore, we see that the manifold constraint estimator that is shown in figure \ref{fig:metric-3d} can determine directly the value of the constraint function from the manifold, allowing the complete evaluation of whether the repeated trajectory is starting to experience deviation. 

\section*{Discussion}

SR has been overlooked in the past because of its known limitations. However, newer advances in component miniaturisation are allowing the exploration of novel approaches to overcome its technical disadvantages. Our sensor array solution proposes a novel approach that challenges the existing single point and physical expansion restrictions by utilising a spectrometer and a light source integrated into a single reflectometer package. Additionally, our proposal provides a linear coverage of 74mm of wafer inspection which is promising for a potential area coverage expansion for the semiconductor and/or R2R manufacturing.

However, the lab-based SR sensor array requires high precision of the angle of incidence, sensor alignment, precise control of the light intensity, sensor integration times, and premium-quality USB devices. All the mentioned requirements represent a limitation of the presented solution as they are a source of variation and potential noise contributors that could affect the sensor's readings. The angle of incidence and sensor alignment in particular are highly dependent on the quality of the 3D-printed parts. Each sensor had to be mechanically adjusted using a Dinolite microscope and a spirit level to ensure proper alignment during the calibration process. This step required at least 30 minutes before attempting the calibration and thus, before starting measuring. This could be solved by modifying the sensor assembly design, incorporating higher-quality moulded parts and adding precise positioning devices in the future.

By contrast, commercially available reflectometers use Halogen and Deuterium stable light sources which cover the full spectrum from UV to Infrared. In this case, for demonstration purposes, we used commercial LEDs which are limited to the Visible (VIS) spectrum (450nm to 700nm). The light intensity of the LEDs was controlled via hardware (HW) and software (SW) . HW-wise, we used potentiometers adjusted to 1.34k$\Omega$ to deliver light intensity of 164Lux (+/-20\%). Software-wise, the integration time was adjusted to 110$\mu$s when measuring the SiO\textsubscript{2} coating to reduce an observed offset vs the uncoated Si reference. Although this is not a widespread practice according to the publicly available manuals \cite{Avantes, ops_manual_filmetrics} and following the National Physical Laboratory (NPL) practices \cite{npl_sensor_2023}, this was the best combination to ensure a good fit of the measured reflectance curve for the stated conditions.

All the aforementioned factors could also be impacted by the quality of the 3D-printed parts. The 3D printer had a tolerance of 0.2mm, which in combination with all the printed parts, could potentially affect the sensor array alignment. The USB hub was also a source of random noise added to the system. The hub presented noise when more than one sensor was connected to it. After Fourier and Lomb-Scargle analysis, it was not possible to locate a frequency to apply a noise reduction filter. Although, in case a frequency could be isolated, adding a filter per pixel would be a high-cost processing, therefore the best noise reduction strategy was to apply a convolution procedure using the Python NumPy library. Knowing the sensor array limitations is key to understanding the development requirements to improve its performance and test its capabilities with flexible substrates. Although there is room for improvement, the results are promising as the sensor array can measure reliable thin-film thickness with less than 2\% thickness variation compared to a well-established SR system when it is positioned in the calibration point, and when there are angle variations below 0.5°. Moreover, the sensor array can detect RMSE and $R^2$ variations when the sample goes 1mm below the calibration point which is a desirable feature for detecting web fluttering failures in R2R manufacturing. Future work with the sensing array includes finding its capability in R2R variable-speed environments, testing with flexible solar cell materials and potential feedback control for the robot arm testing sequence. For the manipulator, the VAE can be used to produce new trajectories that satisfy the desired constraint,but additionally, new constraints can be directly imposed within the Riemannian metric to allow different constraints, such as limits on rotations or positional translation. Furthermore, whilst this work has focused on wafer transportation for inspection through a custom end-effector, in future approaches the sensing array could be attached to the end-effector of the manipulator to allow for R2R inspection with linear motion constraints.

In this research, we presented an automated inspection system using a novel sensor array and a novel robot constraint manipulator. The sensor array is capable of measuring coating thickness less than 2\% error in static and dynamic environments when the wafer remains in the calibration point, and can detect angle variations of 0.5$^\circ$ in the positions described in figure \ref{fig:setup-2}. This can be combined with the robot manipulator to position a wafer to its calibration point whilst adhering to a novel learned constraint manifold, whereby the array can potentially perform surface mapping of the wafer.  

This work sets the footprint for more size-reduced solutions using SR for inspection expansion in R2R systems, and the implementation of SR with more advanced sensor fusion techniques. Within the robotic manipulation field, our method allows for the design of constraint manifolds that are flexible to the constraint that is being imposed whilst maintaining the underlying kinematics of the manipulator. Furthermore, our method allows the direct evaluation of whether the manipulator is deviating from the desired constraint.

\bibliography{sample}

\begin{thebibliography}{10}
\urlstyle{rm}
\expandafter\ifx\csname url\endcsname\relax
  \def\url#1{\texttt{#1}}\fi
\expandafter\ifx\csname urlprefix\endcsname\relax\def\urlprefix{URL }\fi
\expandafter\ifx\csname doiprefix\endcsname\relax\def\doiprefix{DOI: }\fi
\providecommand{\bibinfo}[2]{#2}
\providecommand{\eprint}[2][]{\url{#2}}

\bibitem{r2r_review_2022}
\bibinfo{author}{Maize, K.}, \bibinfo{author}{Mi, Y.}, \bibinfo{author}{Cakmak, M.} \& \bibinfo{author}{Shakouri, A.}
\newblock \bibinfo{journal}{\bibinfo{title}{Real‐time metrology for roll‐to‐roll and advanced inline manufacturing: A review}}.
\newblock {\emph{\JournalTitle{Advanced Materials Technologies}}} \textbf{\bibinfo{volume}{8}}, \doiprefix\url{10.1002/admt.202200173} (\bibinfo{year}{2022}).

\bibitem{inline_imaging_2016}
\bibinfo{author}{Huemer, F.}, \bibinfo{author}{Jamalieh, M.}, \bibinfo{author}{Bammer, F.} \& \bibinfo{author}{Hönig, D.}
\newblock \bibinfo{journal}{\bibinfo{title}{Inline imaging-ellipsometer for printed electronics}}.
\newblock {\emph{\JournalTitle{tm - Technisches Messen}}} \textbf{\bibinfo{volume}{83}}, \bibinfo{pages}{549–556}, \doiprefix\url{10.1515/teme-2015-0067} (\bibinfo{year}{2016}).

\bibitem{inline_thickness_2019}
\bibinfo{author}{Bammer, F.} \& \bibinfo{author}{Huemer, F.}
\newblock \bibinfo{journal}{\bibinfo{title}{Inline thickness measurement with imaging ellipsometry}}.
\newblock {\emph{\JournalTitle{Photonics and Education in Measurement Science 2019}}} \doiprefix\url{10.1117/12.2531940} (\bibinfo{year}{2019}).

\bibitem{inline_atomic_2021}
\bibinfo{author}{Connolly, L.} \& \bibinfo{author}{Cullinan, M.}
\newblock \bibinfo{journal}{\bibinfo{title}{In-line applications of atomic force microscope based topography inspection for emerging roll-to-roll nanomanufacturing processes}}.
\newblock {\emph{\JournalTitle{Metrology, Inspection, and Process Control for Semiconductor Manufacturing XXXV}}} \doiprefix\url{10.1117/12.2584346} (\bibinfo{year}{2021}).

\bibitem{defect_assess_2014}
\bibinfo{author}{Blunt, L.}, \bibinfo{author}{Fleming, L.}, \bibinfo{author}{Elrawemi, M.}, \bibinfo{author}{Robbins, D.} \& \bibinfo{author}{Muhamedsalih, H.}
\newblock \bibinfo{title}{In-line metrology for defect assessment on large area roll 2 roll substrates}.
\newblock In \emph{\bibinfo{booktitle}{Proc. 11th IMEKO TC14 Symposium on Laser Metrology for Precision Measurement and Inspection in Industry}} (\bibinfo{address}{Tsukuba, Japan}, \bibinfo{year}{2014}).

\bibitem{active_optical_2020}
\bibinfo{author}{Marrugo, A.~G.}, \bibinfo{author}{Gao, F.} \& \bibinfo{author}{Zhang, S.}
\newblock \bibinfo{journal}{\bibinfo{title}{State-of-the-art active optical techniques for three-dimensional surface metrology: A review [invited]}}.
\newblock {\emph{\JournalTitle{Journal of the Optical Society of America A}}} \textbf{\bibinfo{volume}{37}}, \doiprefix\url{10.1364/josaa.398644} (\bibinfo{year}{2020}).

\bibitem{vision_film_2020}
\bibinfo{author}{Cho, D.-H.} \emph{et~al.}
\newblock \bibinfo{journal}{\bibinfo{title}{Vision-based high speed wafer film thickness profile estimation with nonlinear regression}}.
\newblock {\emph{\JournalTitle{ODS 2020: Industrial Optical Devices and Systems}}} \doiprefix\url{10.1117/12.2566426} (\bibinfo{year}{2020}).

\bibitem{high_speed_film_2021}
\bibinfo{author}{Cho, D.-H.}, \bibinfo{author}{Park, S.~B.}, \bibinfo{author}{Kim, S.-H.}, \bibinfo{author}{Kim, T.} \& \bibinfo{author}{Lee, K.}
\newblock \bibinfo{journal}{\bibinfo{title}{High-speed wafer film measurement with heterogeneous optical sensor system}}.
\newblock {\emph{\JournalTitle{Metrology, Inspection, and Process Control for Semiconductor Manufacturing XXXV}}} \doiprefix\url{10.1117/12.2584200} (\bibinfo{year}{2021}).

\bibitem{Kim_CNN_2019}
\bibinfo{author}{Kim, M.-G.}
\newblock \bibinfo{journal}{\bibinfo{title}{Improved measurement of thin film thickness in spectroscopic reflectometer using convolutional neural networks}}.
\newblock {\emph{\JournalTitle{International Journal of Precision Engineering and Manufacturing}}} \textbf{\bibinfo{volume}{21}}, \bibinfo{pages}{219–225}, \doiprefix\url{10.1007/s12541-019-00260-4} (\bibinfo{year}{2019}).

\bibitem{hamamatsu_2021}
\bibinfo{author}{Hamamatsu}.
\newblock \emph{\bibinfo{title}{Optical Gauge Series}}.
\newblock \bibinfo{organization}{Hamamatsu} (\bibinfo{year}{2021}).

\bibitem{thickness_eval_2021}
\bibinfo{author}{Grau‐Luque, E.} \emph{et~al.}
\newblock \bibinfo{journal}{\bibinfo{title}{Thickness evaluation of alox barrier layers for encapsulation of flexible pv modules in industrial environments by normal reflectance and machine learning}}.
\newblock {\emph{\JournalTitle{Progress in Photovoltaics: Research and Applications}}} \textbf{\bibinfo{volume}{30}}, \bibinfo{pages}{229–239}, \doiprefix\url{10.1002/pip.3478} (\bibinfo{year}{2021}).

\bibitem{spec_low_cost_2023}
\bibinfo{author}{S\'anchez-Arriaga, N.~E.}, \bibinfo{author}{Tiwari, D.}, \bibinfo{author}{Hutabarat, W.}, \bibinfo{author}{Leyland, A.} \& \bibinfo{author}{Tiwari, A.}
\newblock \bibinfo{journal}{\bibinfo{title}{A spectroscopic reflectance-based low-cost thickness measurement system for thin films: Development and testing}}.
\newblock {\emph{\JournalTitle{Sensors}}} \textbf{\bibinfo{volume}{23}}, \bibinfo{pages}{5326}, \doiprefix\url{10.3390/s23115326} (\bibinfo{year}{2023}).

\bibitem{ind_rob_2023}
\bibinfo{author}{Bogue, R.}
\newblock \bibinfo{journal}{\bibinfo{title}{The role of robots in the electronics industry}}.
\newblock {\emph{\JournalTitle{Industrial Robot: the international journal of robotics research and application}}} \textbf{\bibinfo{volume}{50}}, \bibinfo{pages}{717–721}, \doiprefix\url{10.1108/ir-04-2023-0082} (\bibinfo{year}{2023}).

\bibitem{Stilman_2007}
\bibinfo{author}{Stilman, M.}
\newblock \bibinfo{title}{Task constrained motion planning in robot joint space}.
\newblock In \emph{\bibinfo{booktitle}{2007 IEEE/RSJ International Conference on Intelligent Robots and Systems}}, \bibinfo{pages}{3074–3081}, \doiprefix\url{10.1109/IROS.2007.4399305} (\bibinfo{year}{2007}).

\bibitem{nist_doc_2016}
\bibinfo{author}{O’Connor, A.~C.}, \bibinfo{author}{Beaulieu, T.~J.} \& \bibinfo{author}{Rothrock, G.~D.}
\newblock \bibinfo{journal}{\bibinfo{title}{Economic analysis of technology infrastructure needs for advanced manufacturing: Roll-to-roll manufacturing}}.
\newblock {\emph{\JournalTitle{NIST US Department of Commerce}}} \doiprefix\url{10.6028/nist.gcr.16-008} (\bibinfo{year}{2016}).

\bibitem{Avantes}
\bibinfo{author}{Avantes}.
\newblock \emph{\bibinfo{title}{User Manual For Avantes Software}}.
\newblock \bibinfo{organization}{Avantes} (\bibinfo{year}{2023}).

\bibitem{elmenreich_sensor_fusion}
\bibinfo{author}{Elmenreich, W.}
\newblock \bibinfo{title}{An introduction to sensor fusion}.
\newblock \bibinfo{type}{Tech. Rep.}, \bibinfo{institution}{Vienna University of Technology} (\bibinfo{year}{2002}).

\bibitem{spectro_user_guide_1999}
\bibinfo{author}{Tompkins, H.~G.} \& \bibinfo{author}{McGahan, W.~A.}
\newblock \emph{\bibinfo{title}{Spectroscopic ellipsometry and reflectometry a user’s guide}} (\bibinfo{publisher}{Wiley}, \bibinfo{year}{1999}).

\bibitem{optical_film_book_1991}
\bibinfo{author}{Heavens, O.~S.}
\newblock \emph{\bibinfo{title}{Optical properties of thin solid films}} (\bibinfo{publisher}{Dover}, \bibinfo{year}{1991}).

\bibitem{imaging_situ_2007}
\bibinfo{author}{Urbánek, M.}, \bibinfo{author}{Spousta, J.}, \bibinfo{author}{Běhounek, T.} \& \bibinfo{author}{Šikola, T.}
\newblock \bibinfo{journal}{\bibinfo{title}{Imaging reflectometry in situ}}.
\newblock {\emph{\JournalTitle{Applied Optics}}} \textbf{\bibinfo{volume}{46}}, \bibinfo{pages}{6309}, \doiprefix\url{10.1364/ao.46.006309} (\bibinfo{year}{2007}).

\bibitem{coefficient_2021}
\bibinfo{author}{Chicco, D.}, \bibinfo{author}{Warrens, M.~J.} \& \bibinfo{author}{Jurman, G.}
\newblock \bibinfo{journal}{\bibinfo{title}{The coefficient of determination r-squared is more informative than smape, mae, mape, mse and rmse in regression analysis evaluation}}.
\newblock {\emph{\JournalTitle{PeerJ Computer Science}}} \textbf{\bibinfo{volume}{7}}, \doiprefix\url{10.7717/peerj-cs.623} (\bibinfo{year}{2021}).

\bibitem{ops_manual_filmetrics}
\bibinfo{author}{Filmetrics}.
\newblock \emph{\bibinfo{title}{Operations Manual for the FILMETRICS F20 Thin-Film Analyzer}}.
\newblock \bibinfo{organization}{Filmetrics} (\bibinfo{year}{2023}).

\bibitem{dogleg_1970}
\bibinfo{author}{Powell, M. J.~D.}
\newblock \bibinfo{title}{A hybrid method for nonlinear equations}.
\newblock In \bibinfo{editor}{Rabinowitz, P.} (ed.) \emph{\bibinfo{booktitle}{Numerical Methods for Nonlinear Algebraic Equations}}, \bibinfo{pages}{87--144} (\bibinfo{publisher}{Gordon and Breach}, \bibinfo{year}{1970}).

\bibitem{sensor_data_fusion_Klein}
\bibinfo{author}{Klein, L.~A.}
\newblock \emph{\bibinfo{title}{Sensor and Data Fusion: A Tool for Information Assessment and Decision Making}} (\bibinfo{publisher}{SPIE Press}, \bibinfo{year}{2018}).

\bibitem{hamamatsu_2021_technical_info}
\bibinfo{author}{Hamamatsu}.
\newblock \emph{\bibinfo{title}{Technical Information - Mini spectrometers}}.
\newblock \bibinfo{organization}{Hamamatsu} (\bibinfo{year}{2021}).

\bibitem{remote_sensing_2022}
\bibinfo{author}{Wang, J.}, \bibinfo{author}{Zhao, X.}, \bibinfo{author}{Deuss, K.~E.}, \bibinfo{author}{Cohen, D.~R.} \& \bibinfo{author}{Triantafilis, J.}
\newblock \bibinfo{journal}{\bibinfo{title}{Proximal and remote sensor data fusion for 3d imaging of infertile and acidic soil}}.
\newblock {\emph{\JournalTitle{Geoderma}}} \textbf{\bibinfo{volume}{424}}, \bibinfo{pages}{115972}, \doiprefix\url{https://doi.org/10.1016/j.geoderma.2022.115972} (\bibinfo{year}{2022}).

\bibitem{intro_manifolds_2018}
\bibinfo{author}{Lee, J.~M.}
\newblock \emph{\bibinfo{title}{Introduction to riemannian manifolds}}.
\newblock Graduate Texts in Mathematics (\bibinfo{publisher}{Springer}, \bibinfo{year}{2018}).

\bibitem{wang_deep_robot_skills_2022}
\bibinfo{author}{Wang, W.}, \bibinfo{author}{Saveriano, M.} \& \bibinfo{author}{Abu-Dakka, F.~J.}
\newblock \bibinfo{journal}{\bibinfo{title}{Learning deep robotic skills on riemannian manifolds}}.
\newblock {\emph{\JournalTitle{IEEE Access}}} \textbf{\bibinfo{volume}{10}}, \bibinfo{pages}{114143–114152}, \doiprefix\url{10.1109/ACCESS.2022.3217800} (\bibinfo{year}{2022}).

\bibitem{reactive_vae_2022}
\bibinfo{author}{Beik-Mohammadi, H.}, \bibinfo{author}{Hauberg, S.}, \bibinfo{author}{Arvanitidis, G.}, \bibinfo{author}{Neumann, G.} \& \bibinfo{author}{Rozo, L.}
\newblock \bibinfo{journal}{\bibinfo{title}{Reactive motion generation on learned riemannian manifolds}}.
\newblock {\emph{\JournalTitle{The International Journal of Robotics Research}}} \textbf{\bibinfo{volume}{42}}, \bibinfo{pages}{729–754}, \doiprefix\url{10.1177/02783649231193046} (\bibinfo{year}{2023}).

\bibitem{vae_2013}
\bibinfo{author}{Kingma, D.~P.} \& \bibinfo{author}{Welling, M.}
\newblock \bibinfo{title}{Auto-encoding variational bayes}.
\newblock In \emph{\bibinfo{booktitle}{2014 2nd International Conference on Learning Representations (ICLR)}}, \bibinfo{pages}{14}, \doiprefix\url{10.48550/arXiv.1312.6114} (\bibinfo{address}{Banff, AB, Canada}, \bibinfo{year}{2013}).

\bibitem{yersak_2014}
\bibinfo{author}{Yersak, A.~S.}, \bibinfo{author}{Lee, Y.~C.}, \bibinfo{author}{Spencer, J.~A.} \& \bibinfo{author}{Groner, M.~D.}
\newblock \bibinfo{journal}{\bibinfo{title}{Atmospheric pressure spatial atomic layer deposition web coating with in situ monitoring of film thickness}}.
\newblock {\emph{\JournalTitle{Journal of Vacuum Science \& Technology A: Vacuum Surfaces and Films}}} \textbf{\bibinfo{volume}{32}}, \bibinfo{pages}{1--8}, \doiprefix\url{10.1116/1.4850176} (\bibinfo{year}{2014}).

\bibitem{mag_fac_1997}
\bibinfo{author}{Bishop, C.~M.}, \bibinfo{author}{Svensen, M.} \& \bibinfo{author}{Williams, C. K.~I.}
\newblock \bibinfo{title}{Magnifcation factors for the som and gtm algorithms}.
\newblock In \emph{\bibinfo{booktitle}{Proc. 1997 Workshop on Self-Organizing Maps}} (\bibinfo{address}{Helsinki, Finland}, \bibinfo{year}{1997}).

\bibitem{moveit_2014}
\bibinfo{author}{Coleman, D.}, \bibinfo{author}{Sucan, I.~A.}, \bibinfo{author}{Chitta, S.} \& \bibinfo{author}{Correll, N.}
\newblock \bibinfo{journal}{\bibinfo{title}{Reducing the barrier to entry of complex robotic software: a moveit! case study}}.
\newblock {\emph{\JournalTitle{Journal of Software Engineering for Robotics}}} \textbf{\bibinfo{volume}{5}}, \bibinfo{pages}{3--16}, \doiprefix\url{10.6092/JOSER_2014_05_01_p3} (\bibinfo{year}{2014}).

\bibitem{npl_sensor_2023}
\bibinfo{author}{National{\ }Physical{\ }Laboratory}.
\newblock \emph{\bibinfo{title}{An Introduction to Sensor Validation}} (\bibinfo{publisher}{YouTube}, \bibinfo{year}{2023}).

\end{thebibliography}

\section*{Acknowledgements}

The work of Ethan Canzini was supported by the EPSRC ICASE Award with Airbus, and the work of Nathan John Espley-Plumb was supported by the EPSRC ICASE Award with Siemens UK. The work of Néstor Sánchez was supported by the EPSRC Network Plus in Digitalised Surface Manufacturing: Towards “World’s Best” Processes (EP/S036180/1), EPSRC Intelligent engineering coatings for in-manufacture and in-service monitoring of critical safety products (CoatIN) (EP/T024607/1) and the Consejo Nacional de Ciencia y Tecnolog\'ia (CONACYT, M\'exico) (CVU: 1059795). This work was co-sponsored and supported by the University of Sheffield, Siemens UK and Airbus UK Assembly Technologies and The Henry Royce Institute at the Imperial College of London, receiving funding and support from UKRI EPSRC through the Made Smarter Innovation-Research Centre for Connected Factories under Grant EP/V062123/1 and in part by the Royal Academy of Engineering (RAEng) and Airbus through the Research Chairs and Senior Research Fellowships Scheme under Grant RCSRF1718/5/41. Ashutosh Tiwari is Airbus/RAEng Research Chair in Digitisation for Manufacturing at the University of Sheffield. The authors would like to thank Divya Tiwari (D.T) and Windo Hutabarat (W.H) for their support, supervision, and contribution to prior research that led to this work. We thank ThorLabs for their permission to reproduce the image for the PY005/M base.

\section*{Author contributions statement}

N.S and E.C conceived the hypothesis and the experiments. N.S designed and built the sensor array and E.C created the SW stack for the manipulator platform. N.S and N.P undertook the sensing-based experimentation and N.P and E.C undertook the robotics-based experimentation. M.F and S.P provided support for paper writing and feedback on experiments. A.L provided support on the sensor array experiments and A.T is the PI and main project supervisor. All authors reviewed the manuscript prior to submission. 

\section*{Additional information}

\subsection*{Correspondence}

Correspondence and requests for materials should be addressed to E.C.

\subsection*{Data Availability}

All sensor data generated and analysed are included in this published article and its supplementary information files. The data and code for the constrained robot manipulation is available from the corresponding author on reasonable request.

\subsection*{Competing Interests}

The following authors and collaborators are commercially interested in the sensor array design under the submitted patent application number GB2417203.3: N.S, D.T, W.H, A.L, and A.T.

\end{document}